\documentclass{article}

\PassOptionsToPackage{numbers}{natbib}
\usepackage[preprint]{neurips_2026}

\usepackage[utf8]{inputenc} % allow utf-8 input
\usepackage[T1]{fontenc}    % use 8-bit T1 fonts
\usepackage{hyperref}       % hyperlinks
\usepackage{url}            % simple URL typesetting
\usepackage{booktabs}       % professional-quality tables
\usepackage{amsfonts}       % blackboard math symbols
\usepackage{nicefrac}       % compact symbols for 1/2, etc.
\usepackage{microtype}      % microtypography
\usepackage{xcolor}         % colors
\usepackage{multirow}                  
\usepackage{graphicx}                   
\usepackage{amsmath}                   
\usepackage{amssymb}  

\title{NeighborDiv: Training-free Zero-shot Generalist Graph Anomaly Detection via Neighbor Diversity}

\author{%
  Kaifeng Wei\thanks{Equal contribution} \\
  Netease Yidun AI Lab \\
  Hangzhou, China \\
  \texttt{hzweikaifeng@corp.netease.com} \\
  \And
  Teng Liu$^*$ \\
  School of Software Technology \\
  Zhejiang University \\
  Ningbo, China \\
  \texttt{liuteng2@zju.edu.cn} \\
  \AND
  Liang Dong \\
  School of Advanced Technology \\
  Xi'an Jiaotong-Liverpool University \\
  Suzhou, China \\
  \texttt{brady118899@gmail.com} \\
  \And
  Xiubo Liang\thanks{Corresponding author} \\
  School of Software Technology \\
  Zhejiang University \\
  Ningbo, China \\
  \texttt{xiubo@zju.edu.cn}
  \And
  Yuke Li$^\dagger$ \\
  Netease Yidun AI Lab \\
  Hangzhou, China \\
  \texttt{liyuke@corp.netease.com} \\
}

\begin{document}

\maketitle

\begin{abstract}
Graph Anomaly Detection (GAD) is increasingly shifting to Generalist GAD (GGAD) for cross-domain "one-for-all" detection, but existing GGAD methods predominantly rely on the neighbor consistency principle, falling into the \textbf{Node-to-Neighbor Consistency Paradigm} for anomaly quantification. These methods suffer from complex training pipelines, heavy training data dependency, high computational costs, and unstable cross-domain generalization. To address these limitations, we propose NeighborDiv, a training-free generalist graph anomaly detection framework based on neighbor diversity. Departing from the dominant Node-to-Neighbor Consistency Paradigm, we shift the focus to the \textbf{Neighbor-to-Neighbor Diversity Paradigm}, and uncover that the internal structural dispersion of a node’s neighbor set is a powerful, independently discriminative anomaly signal. We quantify neighbor diversity via the variance of inter-neighbor feature similarities, which captures how a node organizes its local graph environment, and operates independently of conventional node-to-neighbor consistency frameworks. Extensive experiments under two standard GGAD evaluation paradigms show NeighborDiv achieves state-of-the-art performance, with relative gains of 10.25\% in average AUC and 17.78\% in average AP over the second-best baseline under Single-Domain Independent Training (SDIT), and 6.89\%/9.58\% in AUC/AP under Unified Multi-Domain Training (UMDT), respectively. Notably, NeighborDiv yields zero performance volatility across all datasets, eliminating training-set dependency and establishing a lightweight and highly practical GGAD framework.
\end{abstract}

\section{Introduction}

Graph Anomaly Detection (GAD) has emerged as a critical research direction in graph data mining, aiming to identify nodes, edges, or subgraphs that deviate significantly from normal patterns \cite{akoglu2015graph, ma2021comprehensive, pang2021deep, qiao2025deep}. Conventional GAD methods typically follow a "one-model-for-one-dataset" pattern, requiring specialized training or fine-tuning for each target graph \cite{ding2019deep, liu2021anomaly, qiao2023truncated, cai2024towards}. This paradigm incurs high computational costs, relies heavily on sufficient in-domain data, and suffers from poor generalizability when transferred to unseen graphs or diverse domains.

To address these inherent limitations of dataset-specific GAD, Generalist Graph Anomaly Detection (GGAD) has emerged as a promising new paradigm, targeting "one-for-all" anomaly detection across diverse graph domains without target-specific retraining or fine-tuning \cite{liu2024arc, niu2024zero, qiao2025anomalygfm, zhangia}. The core goal of GGAD is to learn universal anomaly patterns from auxiliary datasets, enabling direct zero-shot inference on unseen graphs, thus drastically reducing data dependency and deployment costs in real-world applications. Recent advances in GGAD have explored various technical routes, including in-context learning \cite{liu2024arc}, neighborhood prompt learning \cite{niu2024zero}, graph foundation model-based universal representation learning \cite{qiao2025anomalygfm}, and invariant feature learning for cross-domain adaptation \cite{zhangia}. Despite their effectiveness, a fundamental commonality unifies the vast majority of existing state-of-the-art GGAD methods: they are built on the neighbor consistency principle, rooted in the cross-domain generalizable homophily assumption of graph data. This principle guides these methods to quantify anomaly scores via the consistency degree between a target node and its neighborhood, and all these methods are exclusively constrained within the Node-to-Neighbor Consistency Paradigm \cite{qiao2023truncated, zhang2026gctam, fan2020anomalydae}.

This dominant Node-to-Neighbor Consistency Paradigm, however, suffers from three critical inherent bottlenecks that severely limit the practicality of existing GGAD methods. First, all these methods require complex training pipelines, including pre-training on large auxiliary datasets, prompt tuning, and invariant feature learning, which significantly increase model complexity and computational overhead. Second, their generalization performance remains highly dependent on the quality, scale, and domain diversity of the training data, leading to severe performance degradation when facing unseen domains with large distribution shifts, and limiting their applicability in scenarios where training data is scarce or inaccessible. Third, training-based GGAD methods inevitably suffer from non-negligible performance volatility across different training runs and source dataset selections, making their deployment in high-stakes industrial scenarios (e.g., financial anti-fraud) unreliable.

Motivated by these critical bottlenecks and the universal homophily-based insight in GAD, this paper explores a fundamental, under-explored research question: Can we design a simple yet effective generalist graph anomaly detection algorithm that leverages the homophily principle from a brand-new perspective, without relying on any training process, while completely breaking free from the dominant Node-to-Neighbor Consistency Paradigm? Existing GGAD methods have exhaustively explored the anomaly signal derived from node-to-neighbor consistency, but have largely overlooked a highly discriminative, independent anomaly signal source: the neighbor-to-neighbor diversity within the node’s own neighbor set, i.e., the internal structural dispersion of the neighbor ensemble. Normal nodes in real-world graphs typically conform to natural community structures, and their neighbors form a relatively homogeneous ensemble; in contrast, anomalous nodes often organize their local neighborhoods in an atypical way, either forming an overly uniform neighbor set or connecting to highly heterogeneous groups from multiple disjoint communities. This behavioral pattern of "how a node organizes its local graph environment" cannot be captured by the node-to-neighbor consistency metric, but can be fully characterized by the neighbor-to-neighbor diversity of its local neighborhood.

Based on this core insight, we propose NeighborDiv, a training-free zero-shot generalist graph anomaly detection framework. Departing completely from the dominant Node-to-Neighbor Consistency Paradigm, we shift the core focus of anomaly quantification from "node-to-neighbor consistency matching" to "neighbor-to-neighbor diversity modeling", and formally propose the Neighbor-to-Neighbor Diversity Paradigm for GGAD. Specifically, we define \textbf{Neighbor Diversity} as the variance of pairwise feature similarities among a node’s one-hop neighbors, which directly quantifies the structural dispersion of the neighbor ensemble. This metric operates independently without any residual-based framework, and naturally distinguishes normal nodes with concentrated neighbor diversity from anomalous nodes with significantly deviated diversity distributions. By eliminating pre-training, fine-tuning, and auxiliary data dependency entirely, NeighborDiv achieves zero performance volatility across all datasets, while maintaining strong cross-domain generalization on unseen graphs. Our key contributions are summarized as follows:

\begin{itemize}
\item \textbf{A new paradigm for GGAD.} We identify neighbor set internal dispersion as a discriminative, independently valid anomaly signal for zero-shot cross-domain detection, shifting GGAD’s core focus from the conventional Node-to-Neighbor Consistency Paradigm to our proposed Neighbor-to-Neighbor Diversity Paradigm. To our knowledge, this is the first training-free GGAD framework without dependency on training data.
\item \textbf{Dual-sided neighbor diversity anomaly quantification.}
We propose Neighbor Diversity, a second-order statistic for anomaly quantification that fundamentally differs from first-order metrics in dominant paradigms. It enables robust unified detection of both overly homogeneous and heterogeneous anomalous neighborhoods, with adaptive graph-level calibration free of domain-specific tuning or training.
\item \textbf{SOTA and stable performance.} Extensive experiments under two standard GGAD protocols show NeighborDiv achieves state-of-the-art results, delivering average relative improvements of 10.25\% in AUC and 17.78\% in AP over the strongest learning-based GGAD baseline under SDIT, and 6.89\%/9.58\% (AUC/AP) under UMDT, with zero performance volatility across all datasets.
\end{itemize}

\section{Related Work}
\label{Related Work}

\paragraph{Graph Anomaly Detection.} Graph anomaly detection (GAD) aims to identify graph elements that deviate significantly from normal patterns \cite{ma2021comprehensive, qiao2025deep}. Existing GAD methods are typically categorized into supervised and unsupervised based on supervision information. Supervised GAD treats detection as binary classification, optimized on labeled nodes \cite{tang2023gadbench}. Early methods use GNNs like GCN \cite{kipf2016semi} or GAT \cite{velivckovic2017graph} to learn node representations. Specialized GNNs further enhance GAD performance: BWGNN \cite{tang2022rethinking} and BernNet \cite{he2021bernnet} utilize spectral filters; GHRN \cite{gao2023addressing} focus on heterophily awareness; CAGAD \cite{xiao2024counterfactual} introduces counterfactual data augmentation for detection robustness; while CARE-GNN \cite{dou2020enhancing} employs reinforcement learning for robust aggregation. However, supervised methods heavily rely on labels and generalize poorly across domains. Unsupervised GAD avoids costly annotations via three main branches. Reconstruction-based methods \cite{ding2019deep, fan2020anomalydae} use autoencoders, treating high reconstruction errors as anomalies. Contrastive learning-based methods \cite{liu2021anomaly, duan2023graph} maximize node-to-neighborhood consistency to distinguish anomalies. Auxiliary task-based methods \cite{huang2022hop, qiao2023truncated, chen2024boosting} leverage local homophily or self-supervised pretext tasks for detection. These methods still follow the "one-model-for-one-dataset" pattern, failing to generalize across domains.

\paragraph{Generalist Graph Anomaly Detection.}
Generalist Graph Anomaly Detection (GGAD) has emerged as a "one-for-all" paradigm, aiming for cross-domain detection without target-domain retraining or fine-tuning \cite{liu2024arc, niu2024zero, qiao2025anomalygfm, zhangia}. ARC \cite{liu2024arc} pioneers this field via in-context learning, utilizing feature alignment and a cross-attentive module for few-shot inference. IA-GGAD \cite{zhangia} enhances zero-shot robustness by mitigating feature and structure shifts through invariant and affinity learning. UNPrompt \cite{niu2024zero} employs neighborhood prompt learning and attribute normalization, using attribute predictability as a universal anomaly measure. AnomalyGFM \cite{qiao2025anomalygfm} leverages a graph foundation model to mine shared anomaly patterns through universal representations. However, existing GGAD methods are constrained by the Node-to-Neighbor Consistency Paradigm, heavy training dependency, and unstable cross-domain generalization. We address these with a training-free GGAD framework built on our novel Neighbor-to-Neighbor Diversity Paradigm.

\paragraph{Intra-neighborhood Structural Metrics for Anomaly Detection.}
Characterizing the internal structure of node neighborhoods has a long history in graph mining and anomaly detection. Classical topological measures, including local clustering coefficient \cite{watts1998collective} and recursive ego-network feature extractor ReFeX \cite{henderson2011s}, focus on topological connectivity patterns but cannot capture attribute-level heterogeneity within neighborhoods. For attributed graphs, the seminal AMEN \cite{perozzi2016scalable} performs neighborhood anomaly ranking via ego-anchored attribute decomposition, while Akoglu et al. \cite{akoglu2015graph} provide a comprehensive survey of related neighborhood dispersion indicators. However, existing intra-neighborhood metrics are either topology-only, ego-anchored, or require downstream supervised learning, and none have been validated as training-free, independently discriminative anomaly scorers for zero-shot generalist graph anomaly detection (GGAD).

\section{Method}
\label{headings}

\begin{figure*}[htbp]
  \centering
  \includegraphics[width=\linewidth]{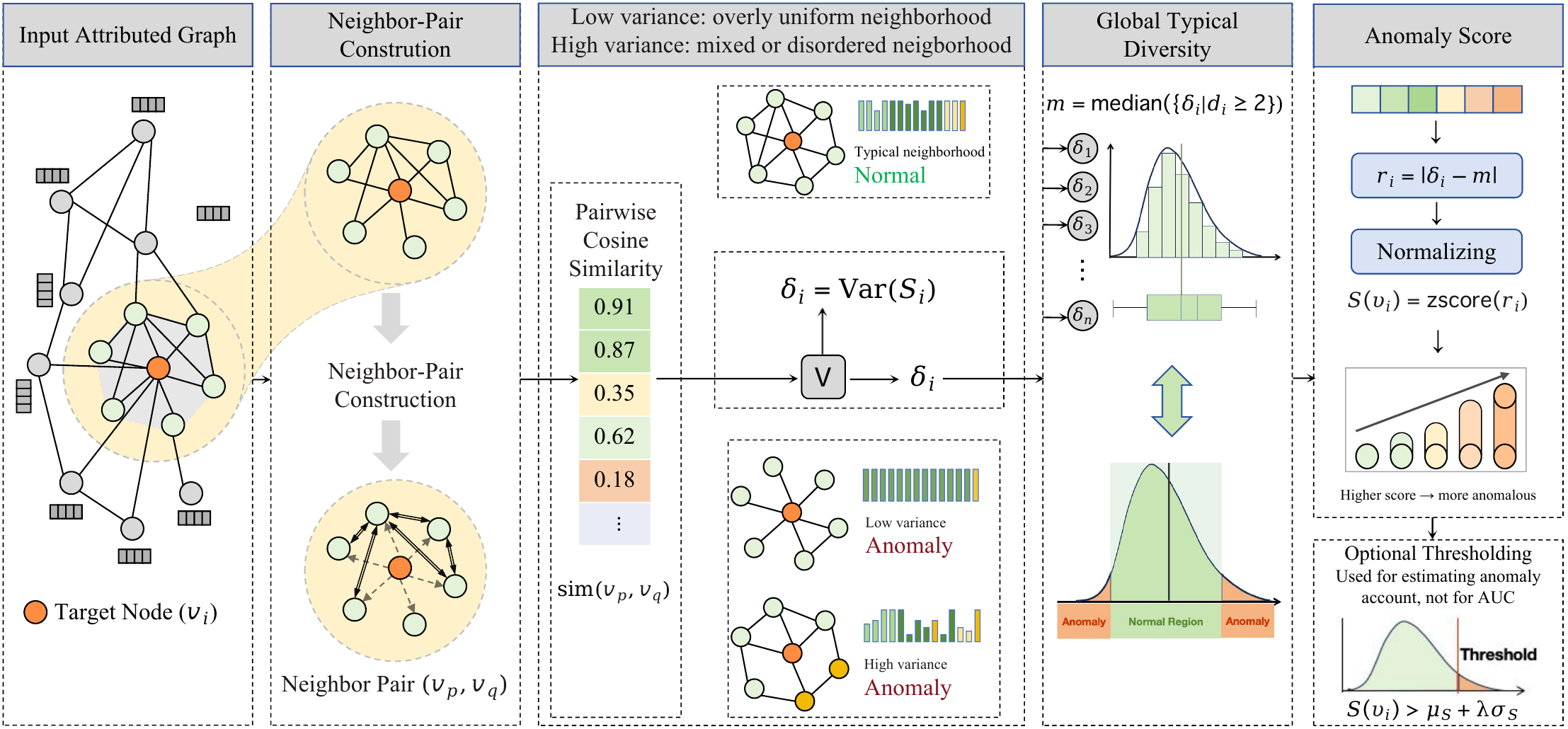}
  \caption{High-level overview of NeighborDiv, our training-free zero-shot generalist graph anomaly detection framework, which quantifies anomalies via the absolute deviation of a node’s neighbor-to-neighbor feature similarity variance from the graph’s median neighborhood diversity level.}
  \label{fig:algorithm_framework}
\end{figure*}

% We propose a \textbf{training-free and unsupervised} graph anomaly detection method based on \textbf{neighbor diversity}. Instead of measuring anomaly by directly comparing a node to a presumed normal prototype, we quantify whether the \textbf{internal organization of its one-hop neighborhood} is atypical. The key intuition is that normal nodes tend to induce relatively stable neighborhood structure, whereas anomalous nodes may correspond to neighborhoods that are unusually homogeneous or heterogeneous. Figure~\ref{fig:algorithm_framework} illustrates the overall framework.
We instantiate our proposed Neighbor-to-Neighbor Diversity Paradigm into the NeighborDiv framework through three steps illustrated in Figure~\ref{fig:algorithm_framework}.
Our method consists of three steps: (1) projecting raw node features into a low-dimensional space for stable similarity estimation; (2) computing a local diversity statistic from pairwise similarities among neighbors; and (3) calibrating this statistic against a graph-level reference to obtain anomaly scores.

\subsection{Low-dimensional Feature Projection}
\label{subsec:feature_projection}
Raw node attributes in attributed graphs may be high-dimensional, noisy,
and inconsistently scaled, which destabilizes pairwise similarity
estimation in the original space. We therefore project them into a
low-dimensional space before measuring neighborhood diversity.

Let $G = (V, E, X)$ be an attributed graph with node set
$V = \{v_1, \dots, v_n\}$, edge set $E$, adjacency matrix
$A \in \{0,1\}^{n \times n}$, and attribute matrix $X \in \mathbb{R}^{n \times d}$.
We perform truncated singular value decomposition (SVD):
\begin{equation}
X = U \Sigma V^\top,
\qquad
X^{(r)} = U_{[:,1:r]} \Sigma_{1:r},
\end{equation}
retaining the top $r$ components ($r{=}8$ in all experiments), and
apply $\ell_1$ row-wise normalization to obtain $\tilde{X}$ with rows
$\tilde{x}_i$. This suppresses noisy/redundant dimensions and makes
pairwise similarity invariant to node-wise scale differences.
% where $r=8$ is fixed in all experiments. Unlike centered PCA, we decompose the raw feature matrix directly to preserve magnitude information in the original attributes. We then apply row-wise normalization to $X^{(r)}$ and denote the resulting representation by $\tilde{X}$, with $\tilde{x}_i$ the representation of node $v_i$. We also conduct an ablation study on this normalization step to evaluate its impact on detection performance.

% This design mainly serves two purposes. First, truncated SVD suppresses noisy and redundant dimensions in the raw feature space. Second, row-wise normalization makes pairwise similarity estimation less sensitive to scale differences across nodes after projection. Although the subsequent cosine similarity is direction-focused, applying SVD before normalization preserves informative feature energy during projection and empirically yields more stable similarity geometry than normalizing the raw feature matrix first. 

\subsection{Neighbor Diversity}
\label{subsec:neighbor_diversity}

For an arbitrary node $v_i$, let its one-hop neighbor set be
\begin{equation}
\mathcal{N}(v_i)=\{v_j \mid A_{ij}=1\},
\end{equation}
and let its degree be
\begin{equation}
d_i = |\mathcal{N}(v_i)|.
\end{equation}

\paragraph{Pairwise similarity.}
Pairwise similarity between neighbors is computed via cosine similarity, which is scale-invariant and focuses on feature directions. For two nodes $v_p$ and $v_q$,
\begin{equation}
\mathrm{sim}(v_p, v_q) = \frac{\tilde{x}_p^\top \tilde{x}_q}{\|\tilde{x}_p\|_2 \|\tilde{x}_q\|_2}.
\label{eq:cosine}
\end{equation}
The $\ell_1$ normalization removes magnitude inconsistency across nodes, while the cosine form further ensures that the similarity depends only on feature directions. In implementation, Eq.~(\ref{eq:cosine}) reduces to an inner product after an additional $\ell_2$ normalization step applied internally to the similarity computation.
% \paragraph{Pairwise similarity.}
% In the normalized feature space, the cosine similarity between two nodes $v_p$ and $v_q$ is defined as
% \begin{equation}
% \mathrm{sim}(v_p,v_q)=
% \frac{\tilde{x}_p^\top \tilde{x}_q}
% {\|\tilde{x}_p\|_2 \|\tilde{x}_q\|_2}.
% \end{equation}

\paragraph{Neighbor diversity by sampled pairwise dispersion.}
If $d_i \ge 2$, the neighbors of $v_i$ form a set of unordered node pairs. A direct implementation would enumerate all $\binom{d_i}{2}$ pairs and compute the variance of their similarities. However, this becomes expensive for high-degree nodes. To make the method scalable, we adopt a sampling-based approximation.

Let
\begin{equation}
\mathcal{P}_i=\{(v_p,v_q)\mid v_p,v_q\in\mathcal{N}(v_i),\; p<q\}
\end{equation}
denote the set of all unordered neighbor pairs, and let
\begin{equation}
M_i = |\mathcal{P}_i| = \binom{d_i}{2}.
\end{equation}
For each node, we uniformly sample at most $k$ pairs from $\mathcal{P}_i$ without replacement:
\begin{equation}
\widetilde{\mathcal{P}}_i \subseteq \mathcal{P}_i,\qquad
|\widetilde{\mathcal{P}}_i| = \min(k, M_i).
\end{equation}
We then collect the sampled pairwise similarities:
\begin{equation}
\widetilde{\mathcal{S}}_i=
\left\{
\mathrm{sim}(v_p,v_q)\mid (v_p,v_q)\in\widetilde{\mathcal{P}}_i
\right\}.
\end{equation}
The \emph{neighbor diversity} of node $v_i$ is defined as the variance of these sampled similarities:
\begin{equation}
D_i = \mathrm{Var}(\widetilde{\mathcal{S}}_i).
\end{equation}

This quantity captures the dispersion of internal relations within the neighborhood of $v_i$. When neighbors are very similar to each other and form an overly uniform local structure, $\widetilde{\mathcal{S}}_i$ becomes concentrated and $D_i$ tends to be small. In contrast, when neighbors belong to several very different local groups, $\widetilde{\mathcal{S}}_i$ becomes more dispersed and $D_i$ tends to be large. Therefore, $D_i$ measures the irregularity of the neighbor-neighbor structure rather than the direct discrepancy between the central node and its neighbors.

\paragraph{Discussion.}
The above sampling procedure reduces per-node cost from $O(d_i^2)$ to
$O(\min(k, d_i^2))$ and empirically serves as a stable Monte Carlo estimator of the full pairwise-similarity variance (theoretically verified in Appendix~\ref{monte_carlo_sampling}, empirically validated in Section~\ref{subsec:ablation_study}). Conceptually, $D_i$ differs from classical
neighborhood descriptors in three respects: it operates in attribute
space rather than edge topology; it is a prototype-free second-order
statistic over neighbor pairs, in contrast to first-order ego-anchored
scores such as AMEN; and the median-calibrated
deviation yields a two-sided score
sensitive to both overly homogeneous and heterogeneous neighborhoods.
Extended analysis and empirical comparison against LCC, NRS, PCD, and
AMEN-Ego are deferred to Appendix~\ref{app:heuristic_baselines} and
Section~\ref{subsec:ablation_study} respectively.

\subsection{Global Calibration and Anomaly Scoring}
\label{subsec:scoring}

A local diversity value $D_i$ alone is not sufficient to determine whether node $v_i$ is anomalous, because the typical diversity level may vary across graphs. We therefore calibrate the local statistic against a graph-level reference.

\paragraph{Graph-level reference.}
We first define the set of valid nodes:
\begin{equation}
\mathcal{V}_{\mathrm{valid}}=\{v_i \mid d_i \ge 2\}.
\end{equation}
Among these nodes, we use the median neighbor diversity as the global reference level:
\begin{equation}
D_{\mathrm{med}}=
\mathrm{Median}
\left(
\{D_i \mid v_i\in\mathcal{V}_{\mathrm{valid}}\}
\right).
\end{equation}
The median is adopted instead of the mean because it is more robust to extreme values in unsupervised settings. We also perform an ablation study on the choice of this global aggregation statistic.

\paragraph{Deviation.}
We then measure how far the neighborhood diversity of each node deviates from this reference:
\begin{equation}
\Delta_i = |D_i - D_{\mathrm{med}}|.
\end{equation}
This absolute deviation is sensitive to both types of atypical neighborhoods: overly homogeneous neighborhoods with unusually small diversity and overly heterogeneous neighborhoods with unusually large diversity.

To make the deviations comparable across nodes, we standardize them on the valid-node set:
\begin{equation}
s_i =
\frac{\Delta_i - \mu_{\Delta}}
{\sigma_{\Delta}}.
\label{eq:zscore}
\end{equation}
where $\mu_{\Delta}$ and $\sigma_{\Delta}$ are the mean and standard deviation of $\Delta_i$ over $\mathcal{V}_{\mathrm{valid}}$, respectively. The resulting $s_i$ is used as the anomaly score of node $v_i$.

\paragraph{Sparse-node handling.}
For nodes with $d_i < 2$, neighbor diversity is undefined because no valid neighbor pair can be formed. We assign such nodes a neutral score $s_i = 0$, which by construction of Eq.~(\ref{eq:zscore}) coincides with the mean of the standardized score distribution over valid nodes. This keeps all anomaly scores on a unified scale and avoids introducing scale-induced bias from a separate scoring function.We further evaluate the sensitivity of this design choice in Appendix~\ref{app:low-degree-handling}.

Notably, the anomaly score does not attempt to estimate whether a node is "normal-like" in an absolute sense. Instead, it measures whether the internal structure induced by that node's neighborhood deviates significantly from the graph's typical neighborhood organization.

\subsection{Binary Decision Rule}
\label{subsec:decision_rule}

When binary anomaly predictions are required, we apply a simple global threshold on the continuous anomaly scores. Let $\mu_s$ and $\sigma_s$ denote the mean and standard deviation of $\{s_i\}_{i=1}^n$, respectively. We define the threshold as
\begin{equation}
\tau = \mu_s + \lambda \sigma_s,
\end{equation}
where $\lambda$ is a hyperparameter.

The final binary prediction is given by
\begin{equation}
\hat{y}_i =
\begin{cases}
1, & \text{if } s_i > \tau,\\
0, & \text{otherwise}.
\end{cases}
\end{equation}
where $\hat{y}_i=1$ indicates that node $v_i$ is predicted as anomalous.

It is worth noting that this thresholding step is only used to estimate the anomalous node set or anomaly count. Ranking-based evaluation metrics such as AUC and AP are determined solely by the continuous anomaly scores $s_i$.

\section{Experiments}
\label{experiments}

\subsection{Experimental Setup}

\paragraph{Datasets.} We use four source training datasets: Facebook \cite{xu2022contrastive}, Amazon \cite{dou2020enhancing}, PubMed \cite{sen2008collective}, and Elliptic \cite{weber2019anti}. For testing, we employ seven target datasets where Cora \cite{mccallum2000automating}, YelpChi \cite{dou2020enhancing}, Reddit \cite{kumar2019predicting},  and T-Finance \cite{tang2022rethinking} are intra-domain samples consistent with training domains, and Tolokers \cite{platonov2023critical}, Disney \cite{sanchez2013statistical}, and Questions \cite{platonov2023critical} are cross-domain samples from unseen domains. Such a domain-aware train-test partition ensures comprehensive and reliable evaluation. Our method is training-free and requires no training data. Detailed dataset statistics are provided in Appendix~\ref{appendix_experimental_setup_dataset}.

\paragraph{Baselines.} We compare with three mainstream paradigms: unsupervised GAD (AnomalyDAE \cite{fan2020anomalydae}, TAM \cite{qiao2023truncated}, GADAM \cite{chen2024boosting}), supervised GAD (GCN \cite{kipf2016semi}, GAT \cite{velivckovic2017graph}, BWGNN \cite{tang2022rethinking}, GHRN \cite{gao2023addressing}), and state-of-the-art zero-shot GGAD methods (AnomalyGFM \cite{qiao2025anomalygfm}, IA-GGAD \cite{zhangia}, UNPrompt \cite{niu2024zero}). Detailed implementations are provided in Appendix~\ref{appendix_experimental_setup_baselines}.

\paragraph{Evaluation Metric.} Following standard practice in graph anomaly detection research \cite{qiao2025anomalygfm, zhangia, niu2024zero, liu2024arc}, we adopt two widely-used evaluation metrics, AUC (Area Under the ROC Curve) and AP (Average Precision), to comprehensively assess the performance of all compared methods. For both metrics, a higher value indicates superior anomaly detection capability.

\paragraph{Implementation Details.}
All experiments are conducted in a fully unsupervised zero-shot setting. We fix the SVD projection dimension at $r=8$ and the threshold coefficient at $\lambda=1.0$. Our method is implemented in PyTorch and requires no training or parameter optimization. Full implementation details are provided in Appendix~\ref{appendix_experimental_setup_implementation_details}.

\subsection{Main Results}
\label{main_results}
%We evaluate NeighborDiv under two standard GGAD protocols: Single-Domain Independent Training (SDIT) and Unified Multi-Domain Training (UMDT), to enable a more thorough and fair comparison.
To reconcile the inconsistent evaluation protocols in the existing GGAD literature and enable a fully fair and unbiased comparison with SOTA baselines, we evaluate NeighborDiv under two unified protocols: Single-Domain Independent Training (SDIT) and Unified Multi-Domain Training (UMDT). Full protocol alignment with prior works is detailed in Appendix~\ref{appendix_evaluation_protocols}.

\paragraph{Single-Domain Independent Training (SDIT).} 
Table~\ref{tab:perf_average} reports average AUC across four source training sets and seven test sets, with aggregated metrics reducing training-set bias for robust generalization ability measurement (per-training-set details in Appendix~\ref{per_dataset_under_sdit}).
Notably, our training-free NeighborDiv yields SOTA average performance, showing relative gains of 10.25\% (AUC) and 17.78\% (AP) over the strongest learning-based baseline IA-GGAD; full AP results are provided in Appendix~\ref{app:main_ap}. Averaging across independent source datasets in SDIT eliminates source domain cherry-picking, and NeighborDiv yields zero performance volatility across all sets.
This superiority is particularly remarkable as it bypasses the expensive training required by other GGAD algorithms, demonstrating that the proposed Neighbor-to-Neighbor Diversity Paradigm captures more fundamental and universal anomaly signals than traditional consistency-based learning models.

The performance gap widens significantly under domain shift. While learning-based baselines (e.g., UNPrompt \cite{niu2024zero} and AnomalyGFM \cite{qiao2025anomalygfm}) perform competitively on intra-domain sets, they suffer catastrophic degradation on cross-domain benchmarks like Tolokers and Questions. Conversely, NeighborDiv maintains stable, high performance across all scenarios. This resilience suggests that baselines overfit to domain-specific features, whereas NeighborDiv utilizes neighbor diversity as a universal structural invariant that is robust to varying feature distributions. Consequently, our method serves as a more reliable generalist for real-world deployments with unknown or unlabeled target domains.

\begin{table*}[t]
    \centering
    \scriptsize
    \setlength{\tabcolsep}{5pt}
    \caption{AUC performance under the SDIT protocol. Each metric represents the mean result across four independent training datasets. (\textbf{bold}: highest value, \underline{underlined}: second highest value)}
    \label{tab:perf_average}
    \begin{tabular}{l|ccccccc|c}
        \toprule
        Method & Cora & YelpChi & Reddit & T-Finance & Tolokers & Disney & Questions & Avg \\
        \midrule
        \multicolumn{9}{c}{\textit{Unsupervised GAD}} \\
        AnomalyDAE (ICASSP'20) & 0.5591 & 0.4864 & 0.5206 & 0.3633 & \underline{0.5170} & 0.4739 & 0.5081 & 0.4898 \\
        TAM (NeurIPS'23)       & \underline{0.7387} & 0.4983 & 0.5565 & 0.5232 & 0.5121 & 0.2359 & 0.5192 & 0.5120 \\
        GADAM (ICLR'24)        & 0.5144 & 0.4844 & 0.5066 & \underline{0.7431} & 0.5074 & 0.4686 & 0.4883 & 0.5304 \\
        \cmidrule(l){1-9}
        \multicolumn{9}{c}{\textit{Supervised GAD}} \\
        GCN (ICLR'17)          & 0.4901 & 0.4258 & \underline{0.5698} & 0.4180 & 0.4921 & 0.5424 & 0.5200 & 0.4940 \\
        GAT (ICLR'18)          & 0.4868 & 0.5260 & 0.5149 & 0.4963 & 0.4777 & 0.4785 & 0.4801 & 0.4943 \\
        BWGNN (ICML'22)        & 0.5013 & 0.4897 & 0.4854 & 0.4879 & 0.4627 & 0.4689 & 0.4679 & 0.4805 \\
        GHRN (WWW'23)          & 0.5099 & 0.4696 & 0.4878 & 0.5855 & 0.4673 & 0.5107 & 0.4427 & 0.4962 \\
        \cmidrule(l){1-9}
        \multicolumn{9}{c}{\textit{Generalist GAD}} \\
        UNPrompt (IJCAI'25)  & 0.6109 & \underline{0.5368} & 0.5463 & 0.2327 & 0.4497 & 0.6385 & 0.4738 & 0.4984 \\        
        AnomalyGFM (KDD'25) & 0.4388 & 0.4694 & 0.5391 & 0.5322 & 0.4740 & \underline{0.6516} & 0.5099 & 0.5164 \\
        IA-GGAD (NeurIPS'25)   & \textbf{0.8808} & 0.5073 & \textbf{0.5877} & 0.5768 & 0.4827 & 0.4612 & \textbf{0.6086} & \underline{0.5864} \\
        % & ARC (NeurIPS'24)       & \underline{0.8660} & \underline{0.5013} & \textbf{0.5932} & 0.7352 & 0.4922 & 0.4582 & 0.6122 & 0.6083 \\
        \textbf{NeighborDiv (ours)}                  & 0.6166 & \textbf{0.6152} & 0.5122 & \textbf{0.9114} & \textbf{0.5624} & \textbf{0.7246} & \underline{0.5832} & \textbf{0.6465} \\
        \midrule
    \end{tabular}
\end{table*}

\paragraph{Source-Domain Dependency Analysis.} Reliance on auxiliary source data remains a fundamental bottleneck for learning-based GGAD methods. As shown in Table~\ref{tab:source_dependency}, baselines exhibit significant performance volatility across different source domains; for instance, the AUC of AnomalyGFM varies by 0.1096 on Tolokers. This instability reveals that existing models based on consistency inadvertently memorize domain distributions during training, failing to isolate truly universal anomaly signals. While the zero variance of NeighborDiv is a natural consequence of its training-free design, its significance lies in the combination of stability and superior accuracy. By leveraging neighbor diversity as a structural invariant, NeighborDiv entirely eliminates the burden of source selection. It provides a deterministic framework that remains robust even in scenarios where learning paradigms falter due to overfitting on specific domains. 

\begin{table*}[t]
    \centering
    \scriptsize
    \setlength{\tabcolsep}{5pt}
    \caption{Source-domain dependency under the SDIT protocol. Each metric represents the standard deviation of performance across four independent training datasets.}
    \label{tab:source_dependency}
    \begin{tabular}{l|lccccccc|c}
        \toprule
        Metric & Method & Cora & YelpChi & Reddit & T-Finance & Tolokers & Disney & Questions & AvgStd \\
        \midrule
        \multirow{4}{*}{AUC}
        & UNPrompt (IJCAI'25)  & 0.0261 & 0.0523 & 0.0176 & 0.0519 & 0.0293 & 0.0596 & 0.0385 & 0.0393 \\
        & AnomalyGFM (KDD'25)  & 0.0369 & 0.0040 & 0.0358 & 0.0959 & 0.1096 & 0.0246 & 0.0165 & 0.0462 \\
        & IA-GGAD (NeurIPS'25) & 0.0026 & 0.0031 & 0.0056 & 0.0448 & 0.0059 & 0.0018 & 0.0113 & 0.0107 \\
        & \textbf{NeighborDiv (ours)}                 & 0.0000 & 0.0000 & 0.0000 & 0.0000 & 0.0000 & 0.0000 & 0.0000 & \textbf{0.0000} \\
        \midrule
        \multirow{4}{*}{AP}
        & UNPrompt (IJCAI'25)  & 0.0077 & 0.0096 & 0.0030 & 0.0019 & 0.0176 & 0.0287 & 0.0037 & 0.0103 \\
        & AnomalyGFM (KDD'25)  & 0.0067 & 0.0010 & 0.0030 & 0.0351 & 0.0582 & 0.0213 & 0.0010 & 0.0180 \\
        & IA-GGAD (NeurIPS'25) & 0.0025 & 0.0009 & 0.0015 & 0.0269 & 0.0035 & 0.0005 & 0.0015 & 0.0053 \\
        & \textbf{NeighborDiv (ours)}                 & 0.0000 & 0.0000 & 0.0000 & 0.0000 & 0.0000 & 0.0000 & 0.0000 & \textbf{0.0000} \\
        \bottomrule
    \end{tabular}
\end{table*}

\paragraph{Unified Multi-Domain Training (UMDT).} We evaluate performance under the UMDT protocol where baselines leverage multi-domain joint training. As shown in Table~\ref{tab:umdt_average}, even against multi-domain jointly trained baselines, NeighborDiv delivers superior average performance, with a 6.89\% higher average AUC than IA-GGAD; the corresponding AP gain is 9.58\% (all relative improvements), with full results in Appendix~\ref{app:main_ap}. Notably, on the complex T-Finance dataset, our method outperforms the second-best model by a large margin (0.9114 AUC). We provide specific diagnostic analyses in Appendix~\ref{app:diagnostic} to confirm the validity of our T-Finance results. While baselines suffer from feature washing and optimization conflicts across disparate datasets, NeighborDiv bypasses these pitfalls entirely due to its training-free nature. This demonstrates that a robust structural invariant, such as neighbor diversity, can outperform features learned through expensive multi-domain optimization, positioning NeighborDiv as a highly efficient yet powerful generalist for graph anomaly detection.

\begin{table*}[t]
    \centering
    \scriptsize
    \setlength{\tabcolsep}{5pt}
    \caption{AUC performance under the UMDT protocol. All learning-based methods are jointly trained on the union of four source domains. (\textbf{bold}: highest value, \underline{underlined}: second highest value)}
    \label{tab:umdt_average}
    \begin{tabular}{l|ccccccc|c}
        \toprule
        Method & Cora & YelpChi & Reddit & T-Finance & Tolokers & Disney & Questions & Avg \\
        \midrule
        % \multirow{15}{*}{AUC}
        \multicolumn{9}{c}{\textit{Unsupervised GAD}} \\
        AnomalyDAE (ICASSP'20) & 0.6016 & 0.4307 & 0.4566 & 0.3096 & 0.5232 & 0.4782 & 0.5361 & 0.4766 \\
        TAM (NeurIPS'23)       & \underline{0.7338} & 0.4902 & 0.5590 & 0.5599 & 0.5095 & 0.2532 & 0.5108 & 0.5166 \\
        GADAM (ICLR'24)        & 0.5111 & 0.4841 & 0.5237 & \underline{0.7828} & \underline{0.5347} & 0.4294 & 0.5066 & 0.5389 \\
        \cmidrule(l){1-9}
        \multicolumn{9}{c}{\textit{Supervised GAD}} \\
        GCN (ICLR'17)          & 0.5705 & 0.4445 & \textbf{0.6217} & 0.5042 & 0.4787 & 0.5323 & 0.5447 & 0.5281 \\
        GAT (ICLR'18)          & 0.4747 & \underline{0.5581} & 0.5856 & 0.5000 & 0.3665 & 0.5000 & 0.5298 & 0.5021 \\
        BWGNN (ICML'22)        & 0.5358 & 0.5012 & 0.5358 & 0.5612 & 0.3890 & 0.4986 & 0.5261 & 0.5068 \\
        GHRN (WWW'23)          & 0.5632 & 0.4724 & 0.5083 & 0.7210 & 0.3809 & 0.5650 & 0.5212 & 0.5332 \\
        \cmidrule(l){1-9}
        \multicolumn{9}{c}{\textit{Generalist GAD}} \\
        UNPrompt (IJCAI'25)  & 0.5519 & 0.5399 & 0.5312 & 0.2781 & 0.4438 & 0.5771 & 0.4626 & 0.4835 \\
        AnomalyGFM (KDD'25) & 0.4430 & 0.4675 & 0.4764 & 0.5411 & 0.4628 & \textbf{0.7472} & 0.5253 & 0.5233 \\
        IA-GGAD (NeurIPS'25)   & \textbf{0.8809} & 0.5207 & \underline{0.5862} & 0.6868 & 0.4770 & 0.4571 & \textbf{0.6251} & \underline{0.6048} \\
        % & ARC (NeurIPS'24)       & \underline{0.8691} & 0.5467 & 0.6028 & 0.7458 & 0.5047 & 0.4738 & 0.6337 &  0.6252 \\
        \textbf{NeighborDiv (ours)}                  & 0.6166 & \textbf{0.6152} & 0.5122 & \textbf{0.9114} & \textbf{0.5624} & \underline{0.7246} & \underline{0.5832} & \textbf{0.6465} \\
        \midrule
    \end{tabular}
\end{table*}

\subsection{Ablation Study}
\label{subsec:ablation_study}
% \begin{figure}[t]
%   \centering
%   \includegraphics[width=\linewidth]{figures/ablation.pdf}
%   \caption{Ablation studies of NeighborDiv.
%     (a) Diversity statistic and graph-level reference; 
%     (b) Comparison with structure-based heuristics;
%     (c) Efficiency-performance trade-off on T-Finances.}
%   \label{fig:ablation}
% \end{figure}
\begin{figure}[t]
  \centering
  \includegraphics[width=\linewidth]{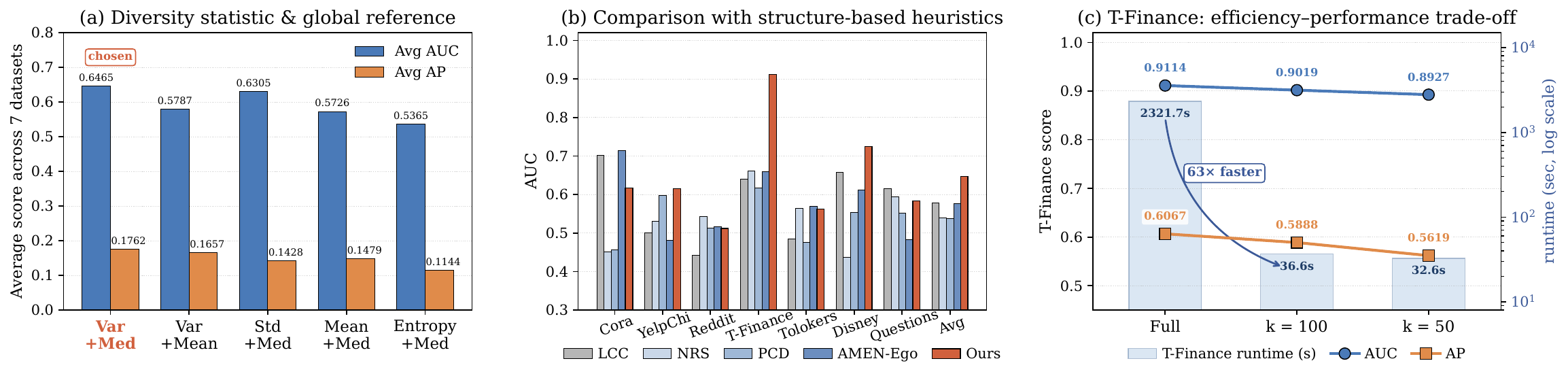}
  \caption{Ablation studies of NeighborDiv.
    (a) Variance with median calibration gives the best overall performance; 
    (b) Neighbor-to-neighbor diversity outperforms structure-based heuristics;
    (c) Neighbor-pair sampling greatly reduces runtime with limited performance loss.}
  \label{fig:ablation}
\end{figure}

We conduct ablation studies to examine three aspects of the proposed method: (1) the choice of diversity statistic and graph-level reference; (2) the distinctiveness of the captured signal relative to structure-based heuristics; and (3) the efficiency-performance trade-off
of the sampling-based approximation. Representative results are summarized in
Figure~\ref{fig:ablation}, with full per-dataset tables deferred to
Appendix~\ref{app:ablation_results}.
\paragraph{Neighborhood diversity and global reference.}
We compare four local diversity statistics (\emph{variance},
\emph{standard deviation}, \emph{mean}, and \emph{entropy}) and, for the variant that uses variance as the local statistic, two choices of graph-level reference
(\emph{median} and \emph{mean}).

Figure~\ref{fig:ablation}(a) reports the average AUC and AP across the seven
test datasets. Among all variants, variance\,+\,median achieves the
highest average AUC (0.6465) and AP (0.1762). Replacing the median with the
mean drops AUC to 0.5787, confirming the importance of robust global
calibration. Standard deviation attains a comparable average AUC (0.6305)
but a substantially worse AP (0.1428), suggesting that variance provides a
more effective signal for ranking anomalous nodes. These results support
variance\,+\,median as the default configuration.

\paragraph{Comparison with structure-based heuristics.}
To verify that Neighbor Diversity captures a signal distinct from
conventional neighborhood characterization approaches, we compare against
four diagnostic baselines representing different paradigms: \textbf{LCC} (local clustering coefficient), a
pure topological cohesion descriptor; \textbf{NRS} (neighborhood residual
score), instantiating the classical node-to-neighbor consistency paradigm
; \textbf{PCD} (propagation consistency decay), a multi-hop cosine-based inconsistency score; and \textbf{AMEN-Ego}, an ego-anchored attributed normality score adapted from AMEN~\cite{perozzi2016scalable} for node-level scoring.
To isolate the effect of the diversity statistic itself, all baselines
are passed through the same graph-level median calibration and z-score
normalization pipeline as our method. Full formal definitions are
provided in Appendix~\ref{app:heuristic_baselines}.

Figure~\ref{fig:ablation}(b) reports the per-dataset AUC. LCC is
competitive on citation-network graphs but fails to generalize across
domains (average AUC 0.5775). NRS and PCD similarly underperform on
average (0.5402 and 0.5378), suggesting that neighbor-to-neighbor
dispersion provides a signal largely complementary to consistency-based
formulations. AMEN-Ego, the most closely related attribute based heuristic baseline(0.5770), still trails our method by 0.0695 AUC (6.95 percentage points); AP results
exhibit the same trend (Appendix~\ref{app:ablation_results}).

\paragraph{Sampling-based approximation and efficiency analysis.}
A direct implementation of neighbor diversity requires evaluating all
unordered neighbor pairs, which can be prohibitive on high degree nodes.
We therefore adopt a sampling-based approximation that uniformly samples
at most $k$ neighbor pairs per node. We compare three settings: Full (all pairs), Sampling ($k{=}100$), and Sampling ($k{=}50$); each sampling variant is averaged over
5 random seeds.

Figure~\ref{fig:ablation}(c) illustrates the efficiency-performance
trade-off on T-Finance, the densest test graph. Runtime drops from 2321.7\,s under exhaustive computation to 36.6\,s with
$k{=}100$, corresponding to a 63$\times$ speedup, while AUC drops only from
0.9114 to 0.9019 and AP from 0.6067 to 0.5888. Across all seven datasets, sampling
preserves performance remarkably well (average AUC 0.6445 for $k{=}100$
and 0.6421 for $k{=}50$, versus 0.6465 for Full), with standard
deviations across random seeds consistently below $1.6\times10^{-3}$ in
AUC, confirming that the Monte Carlo approximation is statistically
stable. These results demonstrate a favorable efficiency-performance
trade-off that is particularly important for dense, large-scale graphs.
Per-dataset results and runtimes are reported in Appendix~\ref{app:ablation_results}.

\paragraph{Additional analyses.}
We verify that $r{=}8$ is the optimal robust default SVD projection dimension for cross-domain zero-shot inference via a sweep over $r\in\{8,16,32,64\}$ (Appendix~\ref{app:svd-rank}), balancing high frequency noise suppression and core structural information retention across all datasets. We further validate our dual-sided detection mechanism via controlled SBM experiments with homophily ratios $h\in\{0.1,0.3,0.5,0.7,0.9\}$, confirming robust detection of both over-homogeneous and over-diverse anomalies across most benchmark datasets’ homophily ranges (Appendix~\ref{app:homophily_sweep}).

\section{Limitations}
\label{tab:limitations}
% The proposed method is training-free and easy to apply across different attributed graphs. However, it still has several limitations. First, it relies on one-hop neighborhood statistics and may therefore miss anomalies that are mainly reflected at higher-order or global structural scales. Second, its effectiveness depends on the quality of node attributes, since neighbor diversity is computed from feature-based similarities in the reduced space. Third, using a single global reference may be suboptimal for highly heterogeneous graphs with diverse local patterns. In addition, the pairwise computation within neighborhoods can become costly for high-degree nodes. Finally, the current formulation is designed for static node anomaly detection and does not explicitly handle dynamic graphs or edge anomalies.
Despite its strong cross-domain performance, NeighborDiv has clear limitations and future research directions: 
(1) Higher-order modeling. Our work focuses on one-hop Neighbor Diversity (standard for node-level GAD); extending to higher-order/global structures could expand detectable anomaly types.
(2) Adaptive references. While a global median provides robust calibration, cluster-aware references could better accommodate extremely heterogeneous, multi-community graphs.
(3) Scalability on ultra-dense graphs. Sampling acceleration already optimizes runtime; advanced techniques (e.g., locality-sensitive hashing) can further support web-scale ultra-large graphs.
(4) Complex graph settings. Adapting our framework from static, homogeneous graphs to dynamic, heterogeneous, or edge-level anomaly detection remains an open challenge.

\section{Conclusion}
This paper introduces NeighborDiv, a training-free framework that shifts Generalist Graph Anomaly Detection (GGAD) from the conventional Node-to-Neighbor Consistency to a new Neighbor-to-Neighbor Diversity Paradigm. By characterizing anomalies through the structural dispersion of inter-neighbor similarities, NeighborDiv provides a lightweight and training-free approach to GAD. Extensive experiments across standard GGAD benchmarks demonstrate that NeighborDiv achieves state-of-the-art performance with zero inference volatility. Our findings reveal that neighbor diversity is a powerful, underexplored anomaly signal, proving that superior cross-domain generalization can be achieved through simple, training-free principles rather than increasingly complex training pipelines.

{
\small  % 9pt字体
\bibliographystyle{plainnat}  % 官方bst，自动生成[1]编号与格式
\bibliography{refs}  % 你的.bib文件名（不含后缀）
}

%%%%%%%%%%%%%%%%%%%%%%%%%%%%%%%%%%%%%%%%%%%%%%%%%%%%%%%%%%%%
\clearpage
\appendix

% \section{Technical appendices and supplementary material}
% Technical appendices with additional results, figures, graphs, and proofs may be submitted with the paper submission before the full submission deadline (see above). You can upload a ZIP file for videos or code, but do not upload a separate PDF file for the appendix. There is no page limit for the technical appendices. 

% Note: Think of the appendix as ``optional reading'' for reviewers. The paper must be able to stand alone without the appendix; for example, adding critical experiments that support the main claims to an appendix is inappropriate. 

\section{Theoretical Analysis and Mathematical Properties}
\label{app:theoretical-analysis}

This appendix provides a rigorous mathematical foundation for the proposed NeighborDiv framework, focusing on its statistical invariance under cross-domain shifts, its bi-directional discriminative power, and the convergence guarantees of the sampling strategy.

\subsection{Notations and Preliminaries}
Throughout this analysis, we follow the notations defined in the main text. Let $G=(V, E, X)$ be an attributed graph with $X \in \mathbb{R}^{n \times d}$. For a node $v_i$, its neighbor set is $\mathcal{N}(v_i)$ with degree $d_i = |\mathcal{N}(v_i)|$. The neighbor-to-neighbor cosine similarity is defined as $sim(v_p, v_q) = \frac{\tilde{x}_p^\top \tilde{x}_q}{\|\tilde{x}_p\|_2 \|\tilde{x}_q\|_2}$, where $\tilde{x}$ denotes the projected features. The neighbor diversity is computed as $D_i = \text{Var}(\{sim(v_p, v_q)\}_{(v_p, v_q) \in P_i})$, where $P_i$ is the set of all neighbor pairs. The final anomaly score is derived from the absolute deviation $\Delta_i = |D_i - D_{\text{med}}|$, where $D_{\text{med}}$ is the graph-wide median of diversity.

\subsection{Statistical Invariance and Stability under Feature-Space Shift}

One of the core motivations of NeighborDiv is its training-free zero-shot
applicability across graph domains. We analyze its behavior under feature-space
shifts while keeping the graph topology and neighbor sets unchanged. Let
\(z_i\in\mathbb{R}^r\) denote the projected feature of node \(v_i\) after
SVD-based preprocessing, and let
\begin{equation}
    u_i =
    \begin{cases}
    z_i/\|z_i\|_2, & z_i\neq 0,\\
    0, & z_i=0.
    \end{cases}
\end{equation}
For a valid node \(v_i\) with \(d_i\ge 2\), let \(P_i\) be the set of unordered
neighbor pairs and define
\begin{equation}
    S_i=\{\langle u_p,u_q\rangle:(v_p,v_q)\in P_i\}, \qquad
    D_i=\operatorname{Var}(S_i).
\end{equation}

NeighborDiv further computes
\begin{equation}
D_{\mathrm{med}}
=
\operatorname{Median}\{D_i:v_i\in V_{\mathrm{valid}}\},
\qquad
\Delta_i=|D_i-D_{\mathrm{med}}|,
\qquad
s_i=\frac{\Delta_i-\mu_\Delta}{\sigma_\Delta},
\end{equation}
where \(\mu_\Delta\) and \(\sigma_\Delta\) are computed over valid nodes.

\paragraph{Proposition 1 (Exact invariance to conformal linear transformations).}
Suppose the projected features are transformed as
\begin{equation}
    z_i'=\alpha O z_i,\qquad
    \alpha>0,\qquad O^\top O=I.
\end{equation}
Then NeighborDiv produces identical neighbor-diversity values, anomaly scores,
and rankings. This follows because normalization gives \(u_i'=Ou_i\), and
orthogonal transformations preserve all inner products:
\begin{equation}
    \langle u_p',u_q'\rangle = \langle Ou_p,Ou_q\rangle = \langle u_p,u_q\rangle .
\end{equation}
Thus \(S_i'=S_i\), \(D_i'=D_i\), \(D_{\mathrm{med}}'=D_{\mathrm{med}}\),
\(\Delta_i'=\Delta_i\), and \(s_i'=s_i\) for all valid nodes. Hence the anomaly
ranking is exactly preserved.

For raw feature transformations \(X'=\alpha XR\), where \(\alpha>0\) and
\(R^\top R=I\), truncated SVD coordinates change only up to a global positive
scaling and a common orthogonal basis transformation within the retained
subspace. Therefore, the same invariance applies to conformal
reparameterizations of the raw feature space.

\paragraph{Proposition 2 (Stability under bounded residual perturbations).}
NeighborDiv is not claimed to be exactly invariant to arbitrary additive
translations. Instead, consider the post-projection residual perturbation
\begin{equation}
    z_i'=z_i+e_i,
    \qquad
    \|e_i\|_2\le \epsilon\|z_i\|_2,
    \qquad
    0\le\epsilon<1.
\end{equation}
Then normalized directions change smoothly, and every pairwise cosine similarity
changes by at most
\begin{equation}
    \delta_\epsilon=\frac{4\epsilon}{1-\epsilon}.
\end{equation}
Since all cosine similarities lie in \([-1,1]\), the induced diversity and
median-centered deviation perturbations satisfy
\begin{equation}
    |D_i'-D_i|\le 4\delta_\epsilon,
    \qquad
    |\Delta_i'-\Delta_i|\le 8\delta_\epsilon.
\end{equation}
Let \(\eta_\epsilon=8\delta_\epsilon\). If
\(\sigma_\Delta\ge\sigma_0>0\) and \(\eta_\epsilon\le\sigma_0/4\), then the
standardized anomaly scores satisfy
\begin{equation}
    |s_i'-s_i|
    \le
    B_\epsilon,
    \qquad
    B_\epsilon
    =
    \frac{4\eta_\epsilon}{\sigma_0}
    +
    \frac{4\eta_\epsilon}{\sigma_0^2}.
\end{equation}
Consequently, for any two nodes \(v_i\) and \(v_j\), if their original score gap
satisfies
\begin{equation}
    |s_i-s_j|>2B_\epsilon,
\end{equation}
then their relative ordering is preserved after perturbation.

The correct theoretical claim is that NeighborDiv is exactly invariant to
positive scaling and orthogonal feature reparameterization, and stable under
bounded post-projection perturbations.

\subsection{Bi-directional Discriminative Power of Second-order Metrics}
We demonstrate that NeighborDiv captures anomalies that are inherently invisible to first-order node-to-neighbor consistency paradigms.

\paragraph{Theorem 1 (Failure of First-order Consistency).} Consider an anomaly node $v_i$ connecting two disjoint communities $C_1$ and $C_2$ with mean features $\mu_1$ and $\mu_2$. If $v_i$'s feature $\tilde{x}_i$ is the weighted average $\pi \mu_1 + (1-\pi) \mu_2$, the first-order residual $R_i = \|\tilde{x}_i - \mathbb{E}[\tilde{x}_{j \in \mathcal{N}(v_i)}]\|$ becomes zero, rendering $v_i$ undetectable. 

\paragraph{Theorem 2 (NeighborDiv Discriminability).} 
\begin{itemize}
    \item \textbf{Over-diverse (Type-D):} For the node in Theorem 1, the inter-neighbor similarity distribution is bi-modal (intra-community pairs $\approx 1$, inter-community pairs $\approx 0$). The variance $D_i$ reaches up to $0.25$ (when $\pi=0.5$), significantly exceeding the $D_{\text{med}}$ of normal nodes (typically $< 0.05$).
    \item \textbf{Over-homogeneous (Type-H):} For malicious clusters (e.g., bot farms), $sim(v_p, v_q) \approx 1$ for all pairs, leading to $D_i \to 0$. The deviation $|0 - D_{\text{med}}|$ successfully flags these nodes given $D_{\text{med}} > 0$ in natural graphs.
\end{itemize}

\subsection{Convergence of Monte Carlo Sampling}
\label{monte_carlo_sampling}
To ensure efficiency on dense graphs, we employ a sampling strategy. We model $D_i$ as a \textbf{U-statistic} of order 2 with the kernel $h(v_p, v_q) = sim(v_p, v_q)$.

\paragraph{Theorem 3 (Exponential Convergence Bound).} Let $\hat{D}_i$ be the estimate from $k$ sampled pairs. For any $\epsilon > 0$, the Hoeffding-type bound for U-statistics \cite{hoeffding1992class} states:
\begin{equation}
P(|\hat{D}_i - D_i| > \epsilon) \leq 2 \exp\left( - \frac{k \epsilon^2}{C} \right),
\end{equation}
where $C=8$ given the range of cosine similarity is $[-1, 1]$. This proves that a fixed sampling rate (e.g., $k=100$) provides a high-confidence estimate regardless of the node's total degree $d_i$, justifying our $O(k)$ local complexity.

\subsection{Signal-to-Noise Enhancement via Truncated SVD}
We assume the attribute matrix $X$ consists of a low-rank structural signal $X_{\text{sig}}$ (community membership) and high-frequency noise $X_{\text{noise}}$. By retaining the top-$r$ ($r=8$) singular values, we project nodes onto the underlying manifold where neighbor organization is most prominent. This filtering process prevents noisy feature dimensions from distorting the neighbor similarity distribution, which explains the robustness observed in our ablation studies.

\subsection{Applicability and Theoretical Boundaries}
Consistent with our Limitations section, the discriminative power of NeighborDiv is subject to:
\begin{enumerate}
    \item \textbf{Homophily Range:} Performance is optimal when the homophily 
    $h \in [0.1, 0.7]$. In extremely homophilous graphs ($h \approx 0.9$), 
    detection of Type-D anomalies degrades substantially, while Type-H anomaly 
    detection remains relatively effective but also weakens compared with moderate 
    homophily regimes.
    % \item \textbf{Homophily Range:} Performance is optimal when the homophily $h \in [0.1, 0.7]$. In extremely homophilous graphs ($h > 0.9$), $D_{\text{med}}$ approaches 0, reducing the detection gap for Type-H anomalies, while Type-D anomaly detection remains fully effective.
    \item \textbf{Degree Constraints:} For nodes with $d_i < 2$, diversity is undefined. Such nodes are assigned a neutral score, as their local organization cannot be statistically measured via second-order metrics.
\end{enumerate}

\section{Details of Experimental Setup}
\label{appendix_experimental_setup}
\subsection{Datasets}
\label{appendix_experimental_setup_dataset}

To ensure reliable, credible experimental results, training and test datasets are selected to maximize diversity in domain, scale, anomaly type, and anomaly rate, aiming to comprehensively evaluate GAD methods’ cross-domain generalization and SOTA models’ training-set dependency. As shown in Table \ref{tab:dataset_statistics}, we summarize the statistics and characteristics of all training and test datasets (Feat. Dim. = Feature Dimension, Ano. Type = Anomaly Type, Ano. Rate = Anomaly Rate).

\begin{itemize}
\item Domain Diversity. The training set covers four distinct GAD domains: social network (Facebook \cite{xu2022contrastive}), e-commerce (Amazon \cite{dou2020enhancing}), citation network (PubMed \cite{sen2008collective}), and finance (Elliptic \cite{weber2019anti}), avoiding single-domain overfitting. The test set includes 7 domains: 4 intra-domain (Cora \cite{mccallum2000automating}, YelpChi \cite{dou2020enhancing}, Reddit \cite{kumar2019predicting}, T-Finance \cite{tang2022rethinking}, corresponding to training domains) for intra-domain generalization, and 3 cross-domain (Tolokers \cite{platonov2023critical}, Disney \cite{sanchez2013statistical}, Questions \cite{platonov2023critical}) to test cross-domain transfer and expose SOTA’s training-set dependency.

\item Scale Diversity. Datasets span small to ultra-large graphs: node count 124 (Disney) to 48,921 (Questions), edge count 335 (Disney) to 21,222,543 (T-Finance), feature dimension 10D (T-Finance) to 1,433D (Cora). This validates method scalability and avoids size bias.

\item Anomaly Diversity. Datasets include both real-world anomalies (e.g., fake reviews, banned users) and injected structural anomalies (e.g., Cora, PubMed), ensuring practical relevance and controlled evaluation. Anomaly rates range from 2.31\% (Facebook) to 21.80\% (Tolokers), verifying robustness to class imbalance.
\end{itemize}

\begin{table*}[t]
    \centering
    \small
    \setlength{\tabcolsep}{4pt}
    \caption{Statistics and characteristics of training/test datasets}
    \begin{tabular}{lccccccc}
        \toprule
        Dataset & Role & Domain & \#Nodes & \#Edges & Feat. Dim. & Ano. Type & Ano. Rate \\
        \midrule
        \textbf{PubMed} & Train & Citation Network & 19,717 & 44,338 & 500 & Injected & 3.04\% \\
        \textbf{Facebook} & Train & Social Network & 1,081 & 55,104 & 576 & Real & 2.31\% \\
        \textbf{Amazon} & Train & E-commerce & 10,244 & 175,608 & 25 & Real & 6.76\% \\
        \textbf{Elliptic} & Train & Finance & 46,564 & 73,248 & 93 & Real & 9.80\% \\
        \midrule
        \textbf{Cora} & Test & Citation Network & 2,708 & 5,429 & 1,433 & Injected & 5.53\% \\
        \textbf{Reddit} & Test & Social Network & 10,984 & 168,016 & 64 & Real & 3.33\% \\
        \textbf{YelpChi} & Test & E-commerce & 23,831 & 49,315 & 32 & Real & 5.10\% \\
        \textbf{T-Finance} & Test & Finance & 39,357 & 21,222,543 & 10 & Real & 4.60\% \\
        \textbf{Tolokers} & Test& Crowdsourcing & 11,758 & 519,000 & 10 & Real & 21.80\% \\
        \textbf{Disney} & Test & Co-purchase & 124 & 335 & 28 & Real & 4.84\% \\
        \textbf{Questions} & Test & Q\&A Platform & 48,921 & 153,540 & 301 & Real & 2.98\% \\
        \bottomrule
    \end{tabular}
    \label{tab:dataset_statistics}
\end{table*}

\subsection{Evaluation Protocols}
\label{appendix_evaluation_protocols}
To ensure fully fair, unbiased, and comprehensive comparison with state-of-the-art GGAD methods, we unify two dominant, previously disjoint training-inference protocols widely used in the field:
\paragraph{Unified Multi-Domain Training (UMDT).} Aligned with the evaluation paradigm of ARC \cite{liu2024arc} and IA-GGAD \cite{zhangia}, all learning-based baselines are jointly trained on the union of the four source datasets, followed by direct zero-shot inference on all seven test datasets without target-domain fine-tuning.
\paragraph{Single-Domain Independent Training (SDIT).} This protocol follows the core single-source training paradigm of UNPrompt \cite{niu2024zero} and AnomalyGFM \cite{qiao2025anomalygfm}, where a model is trained on one single source dataset and directly applied to all test sets via zero-shot inference. To eliminate the randomness of source dataset selection and the common pitfall of cherry-picking optimal training sets in existing GGAD evaluations, we independently run this standard single-source pipeline on each of the four diverse source datasets for all learning-based baselines, and report the average performance across the four independent runs for unbiased, robust comparison.

These two protocols enable a thorough evaluation of cross-domain generalization ability and training-set dependency for all methods, under both multi-domain fusion and single-domain specialization settings. Notably, our training-free NeighborDiv requires no training in either protocol, producing identical deterministic results across all settings with zero performance volatility.

\subsection{Baselines}
\label{appendix_experimental_setup_baselines}
We compare our proposed method NeighborDiv with a comprehensive set of state-of-the-art graph anomaly detection (GAD) approaches, covering three distinct paradigms: unsupervised GAD, supervised GAD, and generalist GAD. For unsupervised GAD, we include three representative methods: AnomalyDAE \cite{fan2020anomalydae}, TAM \cite{qiao2023truncated}, and GADAM \cite{chen2024boosting}. For supervised GAD, we evaluate against four competitive GNN-based baselines: GCN \cite{kipf2016semi}, GAT \cite{velivckovic2017graph}, BWGNN \cite{tang2022rethinking}, and GHRN \cite{gao2023addressing}. For the generalist zero-shot GAD setting, we compare with three leading zero-shot approaches: AnomalyGFM \cite{qiao2025anomalygfm}, IA-GGAD \cite{zhangia} and UNPrompt \cite{niu2024zero}, representing the latest advancements in zero-shot generalist GAD.

\paragraph{Baseline Implementation.}
We evaluate three categories of baselines: unsupervised GAD, supervised GAD, and generalist GAD.

\textit{\textbf{· Unsupervised GAD:}} AnomalyDAE and TAM are implemented using the codebase released by AnomalyGFM. GADAM is adapted from its official repository with a two-stage pretrain-then-inference pipeline to enable cross-dataset zero-shot evaluation.

\textit{\textbf{· Supervised GAD:}} GCN, GAT, and BWGNN are implemented via the AnomalyGFM codebase. GHRN is adapted from its official repository following the same pretrain-then-inference paradigm as GADAM, with truncated SVD applied to align node features across datasets to a unified dimensional space.

\textit{\textbf{· Generalist GAD:}} AnomalyGFM, IA-GGAD, and UNPrompt are evaluated using their respective official codebases without modification. For IA-GGAD, the fusion hyperparameter $\lambda \in [0,1]$ controls the relative contribution of the residual anomaly score and the local affinity score in the final anomaly prediction; we use the optimal $\lambda$ values provided by the official repository for datasets where they are available, and set $\lambda = 0.5$ by default for the remaining datasets, giving equal weight to both signals. For UNPrompt, which is originally proposed as a single-source training method, we strictly follow its original design: under the SDIT protocol (Tables~\ref{tab:perf_compare_facebook} to~\ref{tab:perf_compare_elliptic}), UNPrompt is independently trained on each of the four auxiliary datasets and its average performance across these four source datasets is reported in Table~\ref{tab:perf_average}; under the UMDT protocol (Table~\ref{tab:umdt_average}), since joint multi-source training is not part of UNPrompt's design, we extend UNPrompt to support multi-source joint training, following the exact same data setup used for all competing generalist GAD baselines.

For all adapted methods, models are trained exclusively on source datasets and directly applied to target datasets without any fine-tuning or access to target-domain labels. To enable cross-dataset transfer, node features across all datasets are projected to a unified d-dimensional space via truncated SVD, followed by batch normalization.

\subsection{Implementation Details}
\label{appendix_experimental_setup_implementation_details}
All experiments are conducted in the unsupervised setting. Unless otherwise specified, the SVD dimension is fixed to $r=8$ for all datasets. For the binary decision rule, the threshold coefficient is set to $\lambda=1.0$. For the main results reported in Tables~\ref{tab:perf_average}--\ref{tab:umdt_average}, we use the exhaustive computation over all unordered neighbor pairs (denoted as Full). The sampling-based approximation is only used in the ablation and efficiency analysis in Tables~\ref{tab:ablation_sampling} and~\ref{tab:efficiency_sampling}, where we explicitly vary the neighbor-pair sampling budget $k$.
We adopt Full as the default setting in the main experiments to eliminate approximation effects and present the strongest deterministic version of the proposed method. The sampling-based variant is evaluated separately to study the efficiency-performance trade-off.
We report AUC and AP as the main evaluation metrics. Since both metrics depend only on the ranking of anomaly scores, the threshold coefficient $\lambda$ affects only binary anomaly prediction and does not affect ranking-based evaluation. The method is implemented in Python using PyTorch, NumPy, SciPy, and scikit-learn. As the proposed approach is training-free, it does not involve parameter optimization, model selection, or early stopping.

\section{Supplemental Experiments}
\label{appendix_experiment}

\subsection{Per-dataset AP Results of the Main Tables}
\label{app:main_ap}

For completeness, we report the per-dataset Average Precision (AP) values 
corresponding to the main AUC tables in Section~\ref{main_results}. 
Table~\ref{tab:sdit_ap_appendix} reports AP under the SDIT protocol 
(companion to Table~\ref{tab:perf_average}), and 
Table~\ref{tab:umdt_ap_appendix} reports AP under the UMDT protocol 
(companion to Table~\ref{tab:umdt_average}). Consistent with the AUC 
results in the main text, NeighborDiv achieves the best average AP under 
both protocols, with a 17.78\% improvement under SDIT and a 9.58\% 
improvement under UMDT over the strongest learning-based baseline 
(IA-GGAD).

\begin{table*}[t]
    \centering
    \scriptsize
    \setlength{\tabcolsep}{5pt}
    \caption{AP performance under the SDIT protocol. Each metric represents the mean result across four independent training datasets. (\textbf{bold}: highest value, \underline{underlined}: second highest value)}
    \label{tab:sdit_ap_appendix}
    \begin{tabular}{l|ccccccc|c}
        \toprule
        Method & Cora & YelpChi & Reddit & T-Finance & Tolokers & Disney & Questions & Avg \\
        % \multirow{15}{*}{AP}
        \midrule
        \multicolumn{9}{c}{\textit{Unsupervised GAD}} \\
        AnomalyDAE (ICASSP'20) & 0.0687 & 0.0615 & 0.0360 & 0.0371 & 0.2254 & 0.0836 & 0.0326 & 0.0778 \\
        TAM (NeurIPS'23)       & \underline{0.1977} & 0.0516 & \underline{0.0404} & 0.0521 & \underline{0.2332} & 0.0349 & \underline{0.0420} & 0.0931 \\
        GADAM (ICLR'24)        & 0.0555 & 0.0602 & 0.0333 & \underline{0.4554} & 0.2295 & 0.0892 & 0.0314 & 0.1363 \\
        \cmidrule(l){1-9}
        \multicolumn{9}{c}{\textit{Supervised GAD}} \\
        GCN (ICLR'17)          & 0.0519 & 0.0550 & 0.0386 & 0.0394 & 0.2213 & \underline{0.1176} & 0.0385 & 0.0803 \\
        GAT (ICLR'18)          & 0.0504 & \underline{0.0653} & 0.0340 & 0.0461 & 0.2111 & 0.0520 & 0.0276 & 0.0695 \\
        BWGNN (ICML'22)        & 0.0720 & 0.0523 & 0.0329 & 0.0509 & 0.2098 & 0.0665 & 0.0290 & 0.0734 \\
        GHRN (WWW'23)          & 0.0809 & 0.0519 & 0.0334 & 0.0734 & 0.2078 & 0.1010 & 0.0277 & 0.0823 \\
        \cmidrule(l){1-9}
        \multicolumn{9}{c}{\textit{Generalist GAD}} \\
        UNPrompt (IJCAI'25)  & 0.0866 & 0.0601 & 0.0376 & 0.0275 & 0.2010 & 0.0992 & 0.0298 & 0.0774 \\
        AnomalyGFM (KDD'25) & 0.0500 & 0.0553 & 0.0339 & 0.0766 & 0.2183 & 0.1153 & 0.0301 & 0.0828 \\
        IA-GGAD (NeurIPS'25)   & \textbf{0.4835} & 0.0540 & \textbf{0.0431} & 0.1433 & 0.2173 & 0.0546 & \textbf{0.0512} & \underline{0.1496} \\
        % & ARC (NeurIPS'24)       & \textbf{0.4971} & \underline{0.0537} & \underline{0.0426} & 0.2312 & 0.2200 & 0.0675 & 0.0514 & \underline{0.1662} \\
        \textbf{NeighborDiv (ours)}                  & 0.0684 & \textbf{0.0750} & 0.0356 & \textbf{0.6067} & \textbf{0.2459} & \textbf{0.1617} & 0.0400 & \textbf{0.1762}\\
        \bottomrule
    \end{tabular}
\end{table*}

\begin{table*}[t]
    \centering
    \scriptsize
    \setlength{\tabcolsep}{5pt}
    \caption{AP performance under the UMDT protocol. All learning-based methods are jointly trained on the union of four source domains. (\textbf{bold}: highest value, \underline{underlined}: second highest value)}
    \label{tab:umdt_ap_appendix}
    \begin{tabular}{l|ccccccc|c}
        \toprule
        Method & Cora & YelpChi & Reddit & T-Finance & Tolokers & Disney & Questions & Avg \\
        \midrule
        % \multirow{15}{*}{AP}
        \multicolumn{9}{c}{\textit{Unsupervised GAD}} \\
        AnomalyDAE (ICASSP'20) & 0.0698 & 0.0461 & 0.0283 & 0.0317 & 0.2296 & 0.1155 & 0.0373 & 0.0798 \\
        TAM (NeurIPS'23)       & \underline{0.1829} & 0.0496 & 0.0407 & 0.0598 & 0.2308 & 0.0357 & \underline{0.0400} & 0.0914 \\
        GADAM (ICLR'24)        & 0.0538 & 0.0534 & 0.0359 & \underline{0.4842} & \underline{0.2384} & 0.0812 & 0.0336 & 0.1401 \\
        \cmidrule(l){1-9}
        \multicolumn{9}{c}{\textit{Supervised GAD}} \\
        GCN (ICLR'17)          & 0.0603 & 0.0460 & \textbf{0.0487} & 0.0486 & 0.2148 & 0.1198 & 0.0389 & 0.0825 \\
        GAT (ICLR'18)          & 0.0482 & \textbf{0.0776} & 0.0394 & 0.0458 & 0.1663 & 0.0484 & 0.0308 & 0.0652 \\
        BWGNN (ICML'22)        & 0.0752 & 0.0533 & 0.0345 & 0.0521 & 0.1831 & 0.0619 & 0.0336 & 0.0705 \\
        GHRN (WWW'23)          & 0.0781 & 0.0496 & 0.0323 & 0.1060 & 0.1770 & 0.0808 & 0.0333 & 0.0796 \\
        \cmidrule(l){1-9}
        \multicolumn{9}{c}{\textit{Generalist GAD}} \\
        UNPrompt (IJCAI'25)  & 0.0720 & 0.0583 & 0.0336 & 0.0290 & 0.1995 & 0.0701 & 0.0281 & 0.0701 \\
        AnomalyGFM (KDD'25) & 0.0486 & 0.0552 & 0.0317 & 0.0638 & 0.2042 & \underline{0.1278} & 0.0305 & 0.0803 \\
        IA-GGAD (NeurIPS'25)   & \textbf{0.4860} & 0.0556 & \underline{0.0441} & 0.2212 & 0.2122 & 0.0535 & \textbf{0.0533} & \underline{0.1608} \\
        % & ARC (NeurIPS'24)       & \textbf{0.5029} & 0.0631 & \underline{0.0444} & 0.2448 & 0.2217 & 0.0600 & 0.0545 & \underline{0.1702} \\
        \textbf{NeighborDiv (ours)}                  & 0.0684 & \underline{0.0750} & 0.0356 & \textbf{0.6067} & \textbf{0.2459} & \textbf{0.1617} & \underline{0.0400} & \textbf{0.1762}\\
        \bottomrule
    \end{tabular}
\end{table*}

\subsection{Per-training-set Detailed Results under SDIT}
\label{per_dataset_under_sdit}
This appendix provides detailed experimental results to complement the average performance analysis presented in the main text (Table \ref{tab:perf_average}). Specifically, Tables~\ref{tab:perf_compare_facebook} to~\ref{tab:perf_compare_elliptic} report the detailed anomaly detection performance of all evaluated methods when trained on each of the four auxiliary training sets (Facebook, Amazon, PubMed, and Elliptic) separately, followed by zero-shot inference on the same seven diverse test sets (Cora, YelpChi, Reddit, T-Finance, Tolokers, Disney, and Questions). These test sets cover both intra-domain scenarios (matching the domains of the training sets) and cross-domain scenarios (with distinct domains from the training sets), enabling a comprehensive evaluation of each method’s domain adaptability and stability across different training data distributions.

All tables follow the same structure: the rows present two key evaluation metrics (AUC and AP), which are widely adopted in graph anomaly detection to measure the discriminative ability of anomaly detection methods. The columns correspond to the seven test sets, along with an "Avg" column that calculates the average performance of each method across all test sets when trained on the specific training set. The evaluated methods include three state-of-the-art (SOTA) zero-shot GGAD baselines (AnomalyGFM \cite{qiao2025anomalygfm}, IA-GGAD \cite{zhangia}, UNPrompt \cite{niu2024zero} and our proposed training-free GGAD approach. Bold values in the "Avg" column indicate the best performance under the corresponding metric for that training set.

\begin{table*}[t]
    \centering
    \scriptsize
    \setlength{\tabcolsep}{5pt}
    \caption{Performance comparison of different anomaly detection methods (training set: Facebook) (\textbf{bold}: highest value, \underline{underlined}: second highest value)}
    \label{tab:perf_compare_facebook}
    \begin{tabular}{l|lccccccc|c}
        \toprule
        Metric & Method & Cora & YelpChi & Reddit & T-Finance & Tolokers & Disney & Questions & Avg \\
        \midrule
        \multirow{15}{*}{AUC}
        & \multicolumn{9}{c}{\textit{Unsupervised GAD}} \\
        & AnomalyDAE (ICASSP'20) & 0.4667 & 0.5836 & 0.5450 & 0.4182 & 0.5457 & 0.5531 & 0.4359 & 0.5069 \\
        & TAM (NeurIPS'23)       & \underline{0.7415} & 0.4968 & 0.5546 & 0.5120 & 0.5133 & 0.2312 & 0.5219 & 0.5102 \\
        & GADAM (ICLR'24)        & 0.5267 & 0.4856 & 0.5081 & \underline{0.7631} & 0.4695 & 0.4596 & 0.4879 & 0.5286 \\
        \cmidrule(l){2-10}
        & \multicolumn{9}{c}{\textit{Supervised GAD}} \\
        & GCN (ICLR'17)          & 0.4675 & 0.3954 & 0.5349 & 0.4194 & 0.5051 & 0.5061 & 0.5026 & 0.4759 \\
        & GAT (ICLR'18)          & 0.4579 & \underline{0.5873} & 0.4980 & 0.5072 & \textbf{0.6118} & 0.4579 & 0.4639 & 0.5120 \\
        & BWGNN (ICML'22)        & 0.4359 & 0.4774 & 0.5040 & 0.4854 & \underline{0.5781} & 0.6102 & 0.4423 & 0.5047 \\
        & GHRN (WWW'23)          & 0.4168 & 0.4799 & 0.4524 & 0.5851 & 0.5043 & \underline{0.6696} & 0.4277 & 0.5051 \\
        \cmidrule(l){2-10}
        & \multicolumn{9}{c}{\textit{Generalist GAD}} \\
        & UNPrompt (IJCAI'25)   & 0.6281 & 0.5850 & 0.5366 & 0.2096 & 0.4507 & 0.6155 & 0.4835 & 0.5013 \\
        & AnomalyGFM (KDD'25)   & 0.4879 & 0.4729 & \textbf{0.5820} & 0.6170 & 0.4931 & 0.6362 & 0.5019 & 0.5416 \\
        & IA-GGAD (NeurIPS'25)  & \textbf{0.8784} & 0.5075 & \underline{0.5798} & 0.5613 & 0.4845 & 0.4613 & \textbf{0.6028} & \underline{0.5822} \\
        & \textbf{NeighborDiv (ours)}                  & 0.6166 & \textbf{0.6152} & 0.5122 & \textbf{0.9114} & 0.5624 & \textbf{0.7246} & \underline{0.5832} & \textbf{0.6465} \\
        \midrule
        \multirow{15}{*}{AP}
        & \multicolumn{9}{c}{\textit{Unsupervised GAD}} \\
        & AnomalyDAE (ICASSP'20) & 0.0490 & \underline{0.0804} & 0.0383 & 0.0500 & 0.2378 & 0.0969 & 0.0255 & 0.0826 \\
        & TAM (NeurIPS'23)       & \underline{0.2000} & 0.0508 & \underline{0.0398} & 0.0492 & 0.2339 & 0.0347 & \underline{0.0425} & 0.0930 \\
        & GADAM (ICLR'24)        & 0.0566 & 0.0604 & 0.0330 & \underline{0.4566} & 0.2070 & 0.1017 & 0.0320 & 0.1353 \\
        \cmidrule(l){2-10}
        & \multicolumn{9}{c}{\textit{Supervised GAD}} \\
        & GCN (ICLR'17)          & 0.0489 & 0.0401 & 0.0355 & 0.0385 & 0.2261 & \underline{0.2123} & 0.0380 & 0.0913 \\
        & GAT (ICLR'18)          & 0.0472 & \textbf{0.0824} & 0.0324 & 0.0485 & \underline{0.2668} & 0.0528 & 0.0273 & 0.0796 \\
        & BWGNN (ICML'22)        & 0.0461 & 0.0513 & 0.0327 & 0.0404 & \textbf{0.2705} & 0.1218 & 0.0256 & 0.0841 \\
        & GHRN (WWW'23)          & 0.0486 & 0.0557 & 0.0293 & 0.0521 & 0.2147 & \textbf{0.2329} & 0.0268 & 0.0943 \\
        \cmidrule(l){2-10}
        & \multicolumn{9}{c}{\textit{Generalist GAD}} \\
        & UNPrompt (IJCAI'25)   & 0.0930 & 0.0691 & 0.0348 & 0.0267 & 0.2054 & 0.1058 & 0.0314 & 0.0809 \\
        & AnomalyGFM (KDD'25)   & 0.0591 & 0.0566 & 0.0373 & 0.1236 & 0.2360 & 0.1166 & 0.0304 & 0.0942 \\
        & IA-GGAD (NeurIPS'25)  & \textbf{0.4801} & 0.0541 & \textbf{0.0410} & 0.1348 & 0.2189 & 0.0545 & \textbf{0.0505} & \underline{0.1477} \\
        & \textbf{NeighborDiv (ours)}                  & 0.0684 & 0.0750 & 0.0356 & \textbf{0.6067} & 0.2459 & 0.1617 & 0.0400 & \textbf{0.1762}\\
        \bottomrule
    \end{tabular}
\end{table*}

Table \ref{tab:perf_compare_facebook} presents the detailed results when the training set is Facebook. It can be observed that our method achieves the highest average AUC (0.6465) and average AP (0.1762) among all methods, outperforming the second-best baselines (IA-GGAD with an average AUC of 0.5822) by significant margins. Notably, our method exhibits exceptional performance on the T-Finance test set (AUC = 0.9114, AP = 0.6067), which far surpasses all baselines, demonstrating its strong ability to capture anomaly patterns even in domains different from the Facebook training set.

\begin{table*}[t]
    \centering
    \scriptsize
    \setlength{\tabcolsep}{5pt}
    \caption{Performance comparison of different anomaly detection methods (training set: Amazon) (\textbf{bold}: highest value, \underline{underlined}: second highest value)}
    \label{tab:perf_compare_amazon}
    \begin{tabular}{l|lccccccc|c}
        \toprule
        Metric & Method & Cora & YelpChi & Reddit & T-Finance & Tolokers & Disney & Questions & Avg \\
        \midrule
        \multirow{15}{*}{AUC}
        & \multicolumn{9}{c}{\textit{Unsupervised GAD}} \\
        & AnomalyDAE (ICASSP'20) & 0.4774 & 0.5207 & 0.5098 & 0.3786 & \underline{0.5467} & 0.4291 & 0.4827 & 0.4779 \\
        & TAM (NeurIPS'23) & \underline{0.7427} & 0.5020 & 0.5569 & 0.4893 & 0.5103 & 0.2321 & 0.5196 & 0.5075 \\
        & GADAM (ICLR'24) & 0.5216 & 0.4827 & 0.4946 & \underline{0.8244} & 0.4770 & 0.4607 & 0.4857 & 0.5352 \\
        \cmidrule(l){2-10}
        & \multicolumn{9}{c}{\textit{Supervised GAD}} \\
        & GCN (ICLR'17) & 0.4498 & 0.5618 & \textbf{0.6156} & 0.3330 & 0.5105 & 0.5727 & 0.5113 & 0.5078 \\
        & GAT (ICLR'18) & 0.4753 & 0.5014 & 0.5050 & 0.5016 & 0.5061 & 0.4918 & 0.4551 & 0.4909 \\
        & BWGNN (ICML'22) & 0.5067 & 0.4232 & 0.4273 & 0.2852 & 0.4462 & 0.4336 & 0.4875 & 0.4300 \\
        & GHRN (WWW'23) & 0.5004 & 0.4324 & 0.4431 & 0.4672 & 0.4983 & 0.3929 & 0.4544 & 0.4555 \\
        \cmidrule(l){2-10}
        & \multicolumn{9}{c}{\textit{Generalist GAD}} \\
        & UNPrompt (IJCAI'25) & 0.5844 & \underline{0.5696} & 0.5275 & 0.3098 & 0.4711 & 0.5938 & 0.4169 & 0.4962 \\
        & AnomalyGFM (KDD'25) & 0.4400 & 0.4641 & 0.4944 & 0.5975 & 0.3982 & \underline{0.6808} & 0.4971 & 0.5103 \\
        & IA-GGAD (NeurIPS'25) & \textbf{0.8845} & 0.5108 & \underline{0.5911} & 0.6404 & 0.4868 & 0.4587 & \textbf{0.6255} & \underline{0.5997} \\
        & \textbf{NeighborDiv (ours)}                  & 0.6166 & \textbf{0.6152} & 0.5122 & \textbf{0.9114} & \textbf{0.5624} & \textbf{0.7246} & \underline{0.5832} & \textbf{0.6465}\\
        \midrule
        \multirow{15}{*}{AP}
        & \multicolumn{9}{c}{\textit{Unsupervised GAD}} \\
        & AnomalyDAE (ICASSP'20) & 0.0499 & \underline{0.0741} & 0.0348 & 0.0343 & \underline{0.2389} & 0.0525 & 0.0309 & 0.0736 \\
        & TAM (NeurIPS'23) & \underline{0.2115} & 0.0511 & 0.0407 & 0.0463 & 0.2320 & 0.0347 & \underline{0.0418} & 0.0940 \\
        & GADAM (ICLR'24) & 0.0562 & 0.0609 & 0.0315 & \underline{0.5725} & 0.2163 & 0.0990 & 0.0324 & 0.1527 \\
        \cmidrule(l){2-10}
        & \multicolumn{9}{c}{\textit{Supervised GAD}} \\
        & GCN (ICLR'17) & 0.0458 & \textbf{0.0970} & \textbf{0.0447} & 0.0332 & 0.2278 & \underline{0.1156} & 0.0364 & 0.0858 \\
        & GAT (ICLR'18) & 0.0491 & 0.0591 & 0.0326 & 0.0461 & 0.2174 & 0.0495 & 0.0255 & 0.0685 \\
        & BWGNN (ICML'22) & 0.0863 & 0.0424 & 0.0272 & 0.0298 & 0.2006 & 0.0510 & 0.0318 & 0.0670 \\
        & GHRN (WWW'23) & 0.0681 & 0.0428 & 0.0281 & 0.0394 & 0.2165 & 0.0446 & 0.0292 & 0.0670 \\
        \cmidrule(l){2-10}
        & \multicolumn{9}{c}{\textit{Generalist GAD}} \\
        & UNPrompt (IJCAI'25) & 0.0911 & 0.0654 & 0.0353 & 0.0302 & 0.2171 & 0.0650 & 0.0242 & 0.0755 \\        
        & AnomalyGFM (KDD'25) & 0.0499 & 0.0547 & 0.0300 & 0.0790 & 0.1779 & 0.1102 & 0.0292 & 0.0758 \\
        & IA-GGAD (NeurIPS'25) & \textbf{0.4859} & 0.0538 & \underline{0.0440} & 0.1811 & 0.2193 & 0.0540 & \textbf{0.0534} & \underline{0.1559} \\
        & \textbf{NeighborDiv (ours)}                  & 0.0684 & 0.0750 & 0.0356 & \textbf{0.6067} & \textbf{0.2459} & \textbf{0.1617} & 0.0400 & \textbf{0.1762}\\
        \bottomrule
    \end{tabular}
\end{table*}

Table \ref{tab:perf_compare_amazon} reports the results when training on the Amazon dataset. Consistent with the Facebook training scenario, our method maintains the best average performance (AUC = 0.6465, AP = 0.1762), while baselines show varying degrees of performance fluctuation.

\begin{table*}[t]
    \centering
    \scriptsize
    \setlength{\tabcolsep}{5pt}
    \caption{Performance comparison of different anomaly detection methods (training set: PubMed) (\textbf{bold}: highest value, \underline{underlined}: second highest value)}
    \label{tab:perf_compare_pubmed}
    \begin{tabular}{l|lccccccc|c}
        \toprule
        Metric & Method & Cora & YelpChi & Reddit & T-Finance & Tolokers & Disney & Questions & Avg \\
        \midrule
        \multirow{15}{*}{AUC}
        & \multicolumn{9}{c}{\textit{Unsupervised GAD}} \\
        & AnomalyDAE (ICASSP'20) & 0.6561 & 0.4119 & 0.5094 & 0.3397 & 0.5007 & 0.4395 & 0.5534 & 0.4872 \\
        & TAM (NeurIPS'23) & \underline{0.7342} & 0.4958 & 0.5541 & 0.5462 & 0.5145 & 0.2355 & 0.5155 & 0.5137 \\
        & GADAM (ICLR'24) & 0.5034 & 0.4812 & 0.5104 & \underline{0.6045} & \underline{0.5248} & 0.5006 & 0.5022 & 0.5182 \\
        \cmidrule(l){2-10}
        & \multicolumn{9}{c}{\textit{Supervised GAD}} \\
        & GCN (ICLR'17) & 0.5572 & 0.3158 & 0.5454 & 0.3643 & 0.4763 & 0.5338 & 0.5372 & 0.4757 \\
        & GAT (ICLR'18) & 0.4624 & \underline{0.5217} & 0.5038 & 0.4860 & 0.4390 & 0.4932 & 0.4834 & 0.4842 \\
        & BWGNN (ICML'22) & 0.6589 & 0.4592 & 0.5222 & 0.4441 & 0.4516 & 0.3845 & 0.5167 & 0.4910 \\
        & GHRN (WWW'23) & 0.7149 & 0.3590 & 0.5204 & 0.4960 & 0.5062 & 0.4407 & 0.4735 & 0.5015 \\
        \cmidrule(l){2-10}
        & \multicolumn{9}{c}{\textit{Generalist GAD}} \\
        & UNPrompt (IJCAI'25) & 0.5931 & 0.4687 & \underline{0.5542} & 0.1972 & 0.4691 & 0.6184 & 0.4946 & 0.4850 \\        
        & AnomalyGFM (KDD'25) & 0.3996 & 0.4684 & 0.5420 & 0.4079 & 0.3833 & \underline{0.6624} & 0.5067 & 0.4815 \\
        & IA-GGAD (NeurIPS'25) & \textbf{0.8805} & 0.5033 & \textbf{0.5877} & 0.5695 & 0.4855 & 0.4628 & \textbf{0.6030} & \underline{0.5846} \\
        & \textbf{NeighborDiv (ours)}                  & 0.6166 & \textbf{0.6152} & 0.5122 & \textbf{0.9114} & \textbf{0.5624} & \textbf{0.7246} & \underline{0.5832} & \textbf{0.6465}\\
        \midrule
        \multirow{15}{*}{AP}
        & \multicolumn{9}{c}{\textit{Unsupervised GAD}} \\
        & AnomalyDAE (ICASSP'20) & 0.0832 & 0.0446 & 0.0339 & 0.0323 & 0.2173 & 0.0824 & 0.0358 & 0.0756 \\
        & TAM (NeurIPS'23) & \underline{0.1838} & 0.0528 & \underline{0.0399} & 0.0559 & 0.2345 & 0.0350 & \underline{0.0416} & 0.0919 \\
        & GADAM (ICLR'24) & 0.0543 & \underline{0.0611} & 0.0334 & \underline{0.4449} & \underline{0.2395} & 0.0736 & 0.0303 & 0.1339 \\
        \cmidrule(l){2-10}
        & \multicolumn{9}{c}{\textit{Supervised GAD}} \\
        & GCN (ICLR'17) & 0.0606 & 0.0347 & 0.0342 & 0.0346 & 0.2160 & 0.0687 & 0.0401 & 0.0698 \\
        & GAT (ICLR'18) & 0.0488 & 0.0608 & 0.0333 & 0.0449 & 0.1960 & 0.0547 & 0.0280 & 0.0667 \\
        & BWGNN (ICML'22) & 0.1117 & 0.0458 & 0.0370 & 0.0375 & 0.1985 & 0.0438 & 0.0341 & 0.0726 \\
        & GHRN (WWW'23) & 0.1628 & 0.0363 & 0.0373 & 0.0423 & 0.2306 & 0.0572 & 0.0310 & 0.0854 \\
        \cmidrule(l){2-10}
        & \multicolumn{9}{c}{\textit{Generalist GAD}} \\
        & UNPrompt (IJCAI'25) & 0.0757 & 0.0473 & 0.0393 & 0.0261 & 0.2055 & 0.0922 & 0.0321 & 0.0740 \\        
        & AnomalyGFM (KDD'25) & 0.0431 & 0.0544 & 0.0337 & 0.0404 & 0.1665 & \underline{0.1429} & 0.0293 & 0.0729 \\
        & IA-GGAD (NeurIPS'25) & \textbf{0.4846} & 0.0552 & \textbf{0.0430} & 0.1395 & 0.2188 & 0.0550 & \textbf{0.0502} & \underline{0.1495} \\
        & \textbf{NeighborDiv (ours)}                  & 0.0684 & \textbf{0.0750} & 0.0356 & \textbf{0.6067} & \textbf{0.2459} & \textbf{0.1617} & 0.0400 & \textbf{0.1762}\\
        \bottomrule
    \end{tabular}
\end{table*}

Table \ref{tab:perf_compare_pubmed} shows the experimental results with PubMed as the training set. This training scenario is particularly challenging for baselines, as PubMed belongs to the biomedical domain, which is distinct from most training sets. For instance, UNPrompt’s average AUC drops to 0.4850  and AnomalyGFM’s average AUC decreases to 0.4815 (the lowest in all training scenarios). In contrast, our method still achieves the highest average AUC (0.6465) and AP (0.1762) with stable performance. Notably, our method's AP on T-Finance (0.6067) is nearly 15 times higher than AnomalyGFM's (0.0404), further highlighting its superiority in cross-domain anomaly detection.

\begin{table*}[t]
    \centering
    \scriptsize
    \setlength{\tabcolsep}{5pt}
    \caption{Performance comparison of different anomaly detection methods (training set: Elliptic) (\textbf{bold}: highest value, \underline{underlined}: second highest value)}
    \label{tab:perf_compare_elliptic}
    \begin{tabular}{l|lccccccc|c}
        \toprule
        Metric & Method & Cora & YelpChi & Reddit & T-Finance & Tolokers & Disney & Questions & Avg \\
        \midrule
        \multirow{15}{*}{AUC}
        & \multicolumn{9}{c}{\textit{Unsupervised GAD}} \\
        & AnomalyDAE (ICASSP'20) & 0.6360 & 0.4293 & 0.5181 & 0.3168 & 0.4749 & 0.4740 & 0.5605 & 0.4871 \\
        & TAM (NeurIPS'23) & \underline{0.7362} & 0.4986 & 0.5602 & 0.5451 & 0.5102 & 0.2448 & 0.5197 & 0.5164 \\
        & GADAM (ICLR'24) & 0.5059 & 0.4879 & 0.5134 & 0.7805 & 0.5583 & 0.4534 & 0.4772 & 0.5395 \\
        \cmidrule(l){2-10}
        & \multicolumn{9}{c}{\textit{Supervised GAD}} \\
        & GCN (ICLR'17) & 0.4857 & 0.4300 & \underline{0.5833} & 0.5554 & 0.4765 & 0.5569 & 0.5287 & 0.5166 \\
        & GAT (ICLR'18) & 0.5517 & 0.4934 & 0.5529 & 0.4902 & 0.3537 & 0.4710 & 0.5180 & 0.4901 \\
        & BWGNN (ICML'22) & 0.4035 & \underline{0.5991} & 0.4879 & 0.7369 & 0.3747 & 0.4472 & 0.4251 & 0.4963 \\
        & GHRN (WWW'23) & 0.4075 & 0.6072 & 0.5352 & \underline{0.7937} & 0.3602 & 0.5397 & 0.4153 & 0.5227 \\
        \cmidrule(l){2-10}
        & \multicolumn{9}{c}{\textit{Generalist GAD}} \\
        & UNPrompt (IJCAI'25) & 0.6379 & 0.5238 & 0.5667 & 0.2140 & 0.4079 & \textbf{0.7263} & 0.5000 & 0.5109 \\
        & AnomalyGFM (KDD'25) & 0.4275 & 0.4722 & 0.5380 & 0.5062 & \textbf{0.6213} & 0.6268 & 0.5340 & 0.5323 \\
        & IA-GGAD (NeurIPS'25) & \textbf{0.8797} & 0.5076 & \textbf{0.5923} & 0.5358 & 0.4739 & 0.4621 & \textbf{0.6029} & \underline{0.5792} \\
        & \textbf{NeighborDiv (ours)}                  & 0.6166 & \textbf{0.6152} & 0.5122 & \textbf{0.9114} & \underline{0.5624} & \underline{0.7246} & \underline{0.5832} & \textbf{0.6465}\\
        \midrule
        \multirow{15}{*}{AP}
        & \multicolumn{9}{c}{\textit{Unsupervised GAD}} \\
        & AnomalyDAE (ICASSP'20) & 0.0927 & 0.0468 & 0.0369 & 0.0316 & 0.2077 & 0.1026 & 0.0382 & 0.0795 \\
        & TAM (NeurIPS'23) & \underline{0.1955} & 0.0518 & \underline{0.0413} & 0.0571 & 0.2324 & 0.0353 & \underline{0.0421} & 0.0936 \\
        & GADAM (ICLR'24) & 0.0550 & 0.0582 & 0.0354 & \underline{0.3476} & \underline{0.2550} & 0.0826 & 0.0307 & 0.1235 \\
        \cmidrule(l){2-10}
        & \multicolumn{9}{c}{\textit{Supervised GAD}} \\
        & GCN (ICLR'17) & 0.0522 & 0.0482 & 0.0398 & 0.0514 & 0.2153 & 0.0737 & 0.0396 & 0.0743 \\
        & GAT (ICLR'18) & 0.0566 & 0.0589 & 0.0376 & 0.0447 & 0.1642 & 0.0510 & 0.0294 & 0.0632 \\
        & BWGNN (ICML'22) & 0.0440 & \underline{0.0698} & 0.0348 & 0.0957 & 0.1694 & 0.0495 & 0.0244 & 0.0697 \\
        & GHRN (WWW'23) & 0.0441 & 0.0727 & 0.0388 & 0.1596 & 0.1695 & 0.0693 & 0.0237 & 0.0825 \\
        \cmidrule(l){2-10}
        & \multicolumn{9}{c}{\textit{Generalist GAD}} \\
        & UNPrompt (IJCAI'25) & 0.0865 & 0.0586 & 0.0409 & 0.0268 & 0.1758 & \underline{0.1339} & 0.0313 & 0.0791 \\
        & AnomalyGFM (KDD'25) & 0.0477 & 0.0555 & 0.0347 & 0.0633 & \textbf{0.2928} & 0.0915 & 0.0313 & 0.0881 \\
        & IA-GGAD (NeurIPS'25) & \textbf{0.4832} & 0.0530 & \textbf{0.0445} & 0.1176 & 0.2120 & 0.0549 & \textbf{0.0505} & \underline{0.1451} \\
        & \textbf{NeighborDiv (ours)}                  & 0.0684 & \textbf{0.0750} & 0.0356 & \textbf{0.6067} & 0.2459 & \textbf{0.1617} & 0.0400 & \textbf{0.1762}\\
        \bottomrule
    \end{tabular}
\end{table*}

Table~\ref{tab:perf_compare_elliptic} presents the results when training on the Elliptic dataset (a financial graph dataset). Baselines exhibit varying degrees of performance degradation under this training scenario: UNPrompt's average AUC drops to 0.5109 and IA-GGAD's average AUC decreases to 0.5792. Our method, however, still maintains the best average performance (AUC = 0.6465, AP = 0.1762) with consistent results across all test sets. This further confirms that our training-free design eliminates the dependency on training set distribution and enables robust generalization across diverse domains.

Collectively, the detailed results in these four tables validate the core conclusions of the main text: our proposed training-free GGAD approach not only achieves SOTA average performance across all training scenarios but also maintains zero performance volatility (consistent average AUC and AP across all four training sets), while existing baselines suffer from significant performance fluctuations due to their reliance on training data. These detailed results further demonstrate that our method’s neighbor diversity-based anomaly characterization effectively captures universal anomaly patterns, enabling stable and robust zero-shot anomaly detection across diverse graph domains, regardless of the training set used.

\section{Heuristic Baseline Definitions}
\label{app:heuristic_baselines}

For full reproducibility, we provide the complete specification of the four structure-based heuristic baselines used in Table~\ref{tab:ablation_structure-based}. All baselines receive the same feature preprocessing as our method: truncated SVD projection to $r=8$ dimensions followed by L1 row normalization.

\paragraph{Unified calibration pipeline.}
To isolate the effect of the diversity statistic itself from calibration differences, every baseline (and our method) follows an identical post-processing procedure. Given a raw score $R_i$ computed per node, we define the valid-node set $V_{\mathrm{valid}}$ (specified per method below), compute the global median reference $R_{\mathrm{med}} = \mathrm{Median}(\{R_i : v_i \in V_{\mathrm{valid}}\})$, take the absolute deviation $\Delta_i = |R_i - R_{\mathrm{med}}|$, and standardize via z-score:
\begin{equation}
s_i = \frac{\Delta_i - \mu_\Delta}{\sigma_\Delta},\quad v_i \in V_{\mathrm{valid}},
\end{equation}
where $\mu_\Delta$ and $\sigma_\Delta$ are the mean and standard deviation of $\Delta_i$ over $V_{\mathrm{valid}}$. Nodes outside $V_{\mathrm{valid}}$ are assigned a default score of $s_i = 0$ to ensure a complete anomaly ranking.
% Nodes outside $V_{\mathrm{valid}}$ are assigned the median of $\{s_i : v_i \in V_{\mathrm{valid}}\}$ to ensure a complete anomaly ranking.

\paragraph{LCC (Local Clustering Coefficient).}
For each node $v_i$ with $d_i \geq 2$,
\begin{equation}
\mathrm{LCC}_i = \frac{2\,T_i}{d_i(d_i-1)},\qquad T_i = |\{(v_p, v_q) : v_p, v_q \in \mathcal{N}(v_i),\, A_{pq}=1,\, p<q\}|.
\end{equation}
Valid-node set: $V_{\mathrm{valid}} = \{v_i : d_i \geq 2\}$. This baseline captures pure topological cohesion, ignoring node attributes. We note that the edge-density view of the neighbor-induced subgraph, $|\{(v_p, v_q) \in \mathcal{N}(v_i)^2 : A_{pq}=1,\, p<q\}| / \binom{d_i}{2}$, is numerically equivalent to $\mathrm{LCC}_i$ on undirected unweighted graphs and is therefore not reported as a separate baseline.

\paragraph{NRS (Neighborhood Residual Score).}
Let $\hat{A} = D^{-1/2} A D^{-1/2}$ be the symmetrically normalized adjacency, where the degree is clamped at a minimum of 1 to avoid division by zero for isolated nodes. The aggregated representation is $\bar{x}_i = (\hat{A}\tilde{X})_i$, and the raw score is the L2 norm of the residual:
\begin{equation}
\mathrm{NRS}_i = \| \tilde{x}_i - \bar{x}_i \|_2.
\end{equation}
Valid-node set: $V_{\mathrm{valid}} = \{v_i : d_i \geq 1\}$ (a single neighbor suffices for aggregation). NRS instantiates the classical node-to-neighbor consistency paradigm underlying many reconstruction-based GAD methods.

\paragraph{PCD (Propagation Consistency Decay).}
Starting from $h^{(0)} = \tilde{X}$, we perform $L=3$ rounds of symmetric-normalized propagation:
\begin{equation}
h^{(\ell)} = \hat{A}\, h^{(\ell-1)},\quad \ell = 1, \dots, L.
\end{equation}
The raw score accumulates the cosine-based consistency decay across hops:
\begin{equation}
\mathrm{PCD}_i = \sum_{\ell=1}^{L} w_\ell \cdot \left(1 - \cos\!\left(h_i^{(\ell-1)},\, h_i^{(\ell)}\right)\right),
\end{equation}
with hop weights $(w_1, w_2, w_3) = (1.0, 0.7, 0.5)$, emphasizing closer-hop inconsistency. Valid-node set: $V_{\mathrm{valid}} = \{v_i : d_i \geq 1\}$. PCD captures multi-hop representation drift, complementary to the one-hop residual formulation of NRS.

\paragraph{Conceptual comparison and AMEN-Ego baseline.}
\label{app:amen}

AMEN~\cite{perozzi2016scalable} and our neighbor diversity both operate in the attribute space, but characterize fundamentally different aspects of a neighborhood. AMEN's normality score aggregates per-attribute internal and external contributions around an ego-anchored reference, yielding a \emph{first-order} measure of how well neighbor attributes cohere with the central node. In contrast, $D_i$ is the variance of pairwise similarities \emph{among} neighbors and never references the central node's attributes directly; it is a \emph{second-order} statistic on the distribution of inter-neighbor relations. This difference is not stylistic. Consider two anomaly patterns: (i) a node whose neighbors form a tight cluster in attribute space but whose own attributes lie far outside that cluster, and (ii) a node whose neighbors are drawn from several disjoint attribute groups regardless of where the node itself lies. AMEN strongly penalizes case (i), because the internal and external attribute decomposition detects the mismatch between the ego node and its neighbors. However, it has no direct signal for case (ii), since each neighbor may individually align with the ego node. $D_i$ shows the opposite behavior. Case (ii) produces a large variance among pairwise neighbor similarities and is therefore assigned a high anomaly score. Case (i) produces low variance, but can still be identified through the two-sided calibration against $D_{\mathrm{med}}$. Real-world anomalies frequently fall into case (ii). For example, fraudulent accounts may interact with a mixture of otherwise unrelated users, which helps explain the empirical gap on T-Finance. (Table~\ref{tab:ablation_structure-based}).

To make this comparison concrete, we now detail the AMEN-Ego baseline used in Table~\ref{tab:ablation_structure-based}. AMEN was originally proposed as a neighborhood-level scoring method that ranks a predefined set of candidate neighborhoods (e.g., ground-truth social circles) by their attributed normality. Since our evaluation is node-level and no ground-truth community structure is available across our seven benchmark datasets, we adopt the ego-network instantiation used in AMEN's own experiments on datasets lacking ground-truth circles: for each node $v_i$, we take $C_i = \{v_i\} \cup \mathcal{N}(v_i)$ as the candidate neighborhood and $B_i = \{v_b \notin C_i \mid \exists v_p \in C_i,\; A_{pb}=1\}$ as its boundary.

For $\tilde{X}$ linearly rescaled per dimension to $[0,1]$ as required by the size-invariant form of AMEN~\cite{perozzi2016scalable}, the per-attribute internal and external contribution vectors are computed as
\begin{equation}
x_I(f) = \sum_{p,q \in C_i}\!\!\left(A_{pq}-\tfrac{k_p k_q}{2m}\right)\!\tilde{x}_p(f)\tilde{x}_q(f),\quad
x_E(f) = -\!\!\!\!\!\sum_{\substack{(p,b)\in E \\ p\in C_i,\, b\in B_i}}\!\!\!\!\!\left(1 - \min(1,\tfrac{k_p k_b}{2m})\right)\!\tilde{x}_p(f)\tilde{x}_b(f).
\end{equation}
Applying the size-invariant rescaling of AMEN~\cite{perozzi2016scalable} to obtain $\hat{x}_I \in [0,1]^d$ and $\hat{x}_E \in [-1,0]^d$, the $L_2$ closed-form optimal normality is
\begin{equation}
\mathcal{N}^*(C_i) = \|x_+\|_2,\quad \text{where } x = \hat{x}_I + \hat{x}_E,\; x_+ = \max(x, 0).
\end{equation}
When $x$ has no positive entry, we set $\mathcal{N}^*(C_i) = \max_f x(f) \le 0$, matching AMEN's original convention. The raw anomaly statistic is $R_i = -\mathcal{N}^*(C_i)$, so that low normality corresponds to high anomaly, which is then passed through the same median-calibrated z-score pipeline as all other heuristics. The valid-node set, sparse-node backoff, and hyperparameter choices follow the unified protocol described at the start of this appendix.

\section{Full Per-Dataset Ablation Results}
\label{app:ablation_results}

\subsection{Diversity Statistic \& Global Reference}
Table~\ref{tab:ablation_metric} isolates the two design choices of 
Section~\ref{subsec:ablation_study}: the local diversity statistic and the global reference.

\textbf{Median is a more robust reference than mean.} Holding the variance statistic fixed, replacing the median reference with the mean reference decreases the average AUC from $0.6465$ to $0.5787$ and the average AP from $0.1762$ to $0.1657$.
The drop is more evident on datasets whose diversity values contain pronounced tail values, such as Cora and Questions, while it is much smaller on T-Finance. This is consistent with the intuition from robust estimation: the median remains stable under a small fraction of extreme $D_i$ values, whereas the mean is pulled toward these extremes and may shift the reference level in the same direction as the anomalous tail, reducing the contrast between normal and anomalous neighborhoods.

\textbf{Second-order dispersion is the right statistic.} Among the four statistics 
under median calibration, variance attains the best average AUC (0.6465) and AP 
(0.1762). Standard deviation, a monotone transform of variance, yields comparable AUC 
(0.6305) but markedly worse AP (0.1428, $-19\%$ relative): squaring amplifies large 
deviations and concentrates anomalous nodes at the top of the ranking, which AP 
rewards but AUC does not. The first-order mean of pairwise similarities loses all 
dispersion information and drops to 0.5726 AUC. Entropy is worst on average (0.5365), 
as binning discards the continuous shape of the similarity distribution that variance 
preserves. Variance$+$median dominates on 4 of 7 datasets and both averages, and is 
adopted as the default.
\begin{table*}[t]
    \centering
    \caption{Ablation on neighborhood diversity statistics and graph-level reference. The best average results are shown in bold.}
    \label{tab:ablation_metric}
    \resizebox{\textwidth}{!}{
    \begin{tabular}{c|cccccccccc}
    \toprule
    Metric & Diversity Metric & Global Reference & Cora & YelpChi & Reddit & T-Finance & Tolokers & Disney & Questions & Avg \\
    \midrule
    \multirow{5}{*}{AUC}
      & Variance  & Median & 0.6166 & 0.6152 & 0.5122 & 0.9114 & 0.5624 & 0.7246 & 0.5832 & \textbf{0.6465} \\
      & Variance  & Mean   & 0.3215 & 0.6055 & 0.4986 & 0.9078 & 0.4902 & 0.7034 & 0.5241 & 0.5787 \\
      & Std. Dev. & Median & 0.5983 & 0.5648 & 0.4753 & 0.8953 & 0.5643 & 0.7119 & 0.6037 & 0.6305 \\
      & Mean      & Median & 0.4226 & 0.5823 & 0.4733 & 0.9132 & 0.5220 & 0.6575 & 0.4375 & 0.5726 \\
      & Entropy   & Median & 0.7415 & 0.5162 & 0.4581 & 0.4649 & 0.5343 & 0.4350 & 0.6055 & 0.5365 \\
    \midrule
    \multirow{5}{*}{AP}
      & Variance  & Median & 0.0684 & 0.0750 & 0.0356 & 0.6067 & 0.2459 & 0.1617 & 0.0400 & \textbf{0.1762} \\
      & Variance  & Mean   & 0.0398 & 0.0693 & 0.0343 & 0.6031 & 0.2216 & 0.1562 & 0.0356 & 0.1657 \\
      & Std. Dev. & Median & 0.0659 & 0.0648 & 0.0304 & 0.3898 & 0.2467 & 0.1587 & 0.0431 & 0.1428 \\
      & Mean      & Median & 0.0494 & 0.0663 & 0.0315 & 0.5621 & 0.2212 & 0.0782 & 0.0265 & 0.1479 \\
      & Entropy   & Median & 0.2930 & 0.0594 & 0.0291 & 0.0391 & 0.2750 & 0.0511 & 0.0544 & 0.1144 \\
    \bottomrule
    \end{tabular}
    }
\end{table*}

\subsection{Comparison with Structure-based Heuristics}
Table~\ref{tab:ablation_structure-based} compares Neighbor Diversity against four 
structure-based heuristics, all passed through the same median calibration and 
z-score pipeline (Appendix~\ref{app:heuristic_baselines}) to isolate the effect of the 
underlying statistic. Our method achieves the best average AUC (0.6465) and AP 
(0.1762), with the largest per-dataset gain on T-Finance (0.9114 vs.\ 0.6610 for 
the best heuristic), precisely the \emph{over-diverse} regime that second-order 
dispersion is designed to capture, where fraudulent accounts interact with 
heterogeneous user groups. Consistency-based heuristics (NRS 0.5402, PCD 0.5378) 
trail all attribute-aware methods on average, supporting the main claim that 
neighbor-to-neighbor dispersion carries a signal unavailable to first-order 
node-to-neighbor formulations.

\textbf{Dataset-level comparison.} LCC (0.7020) and AMEN-Ego (0.7149) 
both outperform our method on Cora (0.6166), and LCC also leads on Questions 
(0.6154 vs.\ 0.5832); NRS narrowly beats our method on Reddit (0.5435 vs.\ 0.5122). 
These are not contradictions to our claim: Cora uses injected structural anomalies 
whose detectability is driven by topological cohesion and ego-attribute deviation, 
which these heuristics capture by construction. To check that the average gap is 
not an artifact of a single strong dataset, we recompute the AUC average excluding 
T-Finance: our method still averages 0.6024 across the remaining six datasets, 
ahead of LCC (0.5671) and AMEN-Ego (0.5631) by 3.5-3.9 percentage points in AUC. Under AP, the 
gap widens further (0.1762 vs.\ 0.0988 for the closest baseline), reflecting that 
Neighbor Diversity concentrates anomalous nodes at the top of the ranking more 
effectively than any heuristic we evaluate.
\begin{table}[h]
    \centering
    \caption{Comparison with structure-based heuristic baselines. 
    Bold indicates best average performance.}
    \label{tab:ablation_structure-based}
    \resizebox{\textwidth}{!}{
    \begin{tabular}{l|ccccccccc}
    \toprule
    Metric & Method & Cora & YelpChi & Reddit & T-Finance & Tolokers & Disney 
    & Questions & Avg \\
    \midrule
    \multirow{5}{*}{AUC}
    & LCC      & 0.7020 & 0.5000 & 0.4432 & 0.6400 & 0.4842 & 0.6575 & 0.6154 & 0.5775 \\
    & NRS      & 0.4509 & 0.5303 & 0.5435 & 0.6610 & 0.5647 & 0.4364 & 0.5947 & 0.5402 \\
    & PCD      & 0.4567 & 0.5968 & 0.5135 & 0.6167 & 0.4763 & 0.5537 & 0.5511 & 0.5378 \\
    & AMEN-Ego & 0.7149 & 0.4818 & 0.5169 & 0.6602 & 0.5697 & 0.6123 & 0.4832 & 0.5770 \\
    & \textbf{NeighborDiv (ours)}     & 0.6166 & 0.6152 & 0.5122 & 0.9114 & 0.5624 & 0.7246 & 0.5832 & \textbf{0.6465} \\
    \midrule
    \multirow{5}{*}{AP}
    & LCC      & 0.1082 & 0.0511 & 0.0282 & 0.0718 & 0.2088 & 0.1020 & 0.0541 & 0.0892 \\
    & NRS      & 0.0509 & 0.0581 & 0.0356 & 0.0719 & 0.2556 & 0.0628 & 0.0411 & 0.0823 \\
    & PCD      & 0.0497 & 0.0744 & 0.0348 & 0.0731 & 0.2126 & 0.0672 & 0.0389 & 0.0787 \\
    & AMEN-Ego & 0.1477 & 0.0452 & 0.0372 & 0.0751 & 0.2676 & 0.0875 & 0.0313 & 0.0988 \\
    & \textbf{NeighborDiv (ours)}     & 0.0684 & 0.0750 & 0.0356 & 0.6067 & 0.2459 & 0.1617 & 0.0400 & \textbf{0.1762} \\
    \bottomrule
    \end{tabular}
    }
\end{table}

\subsection{Sampling Approximation and Runtime Analysis}
Tables~\ref{tab:ablation_sampling} and~\ref{tab:efficiency_sampling} quantify the 
trade-off between Monte Carlo approximation and exhaustive pairwise enumeration.

\textbf{Performance is preserved across density regimes.} On the seven dataset average, sampling with $k=100$ retains more than $99.6\%$ of the Full AUC 
($0.6445$ vs.\ $0.6465$), while sampling with $k=50$ still retains more than $99.3\%$ 
($0.6421$ vs.\ $0.6465$).
Both sampling budgets also preserve more than $96\%$ of the Full AP. Cora, YelpChi, Disney, and Questions are unaffected, because most nodes satisfy 
$\binom{d_i}{2} \le k$ and the sampler returns all pairs exactly. The gap emerges 
only on denser graphs, with T-Finance as the worst case: AUC 
0.9114$\to$0.9019$\to$0.8927 and AP 0.6067$\to$0.5888$\to$0.5619 for 
Full$\to$$k{=}100$$\to$$k{=}50$. Standard deviations across five seeds remain below 
$1.6\times10^{-3}$ in AUC at every entry, confirming the estimator is stable in 
practice.

\textbf{Runtime scales with graph density.} The speedup is concentrated where it 
matters. On T-Finance, Full enumeration takes 2321.7\,s while $k{=}100$ completes 
in 36.6\,s ($63\times$). Tolokers and Questions see $2{-}3\times$ speedups, and the 
sparsest graphs (Cora, YelpChi, Disney) finish under 4\,s in every configuration, on Cora the sampled variant is marginally slower (3.29\,s vs.\ 3.23\,s) because the 
sampling overhead exceeds the negligible saving when $d_i$ is already small. This 
makes $k{=}100$ a pragmatic default: negligible computational overhead on sparse graphs, and the 
only feasible option on web-scale dense graphs where exhaustive enumeration is 
prohibitive.
\begin{table*}[t]
    \centering
    \caption{Ablation on the sampling-based approximation for neighbor diversity. Full denotes exhaustive computation over all unordered neighbor pairs.}
    \label{tab:ablation_sampling}
    \resizebox{\textwidth}{!}{
    \begin{tabular}{c|ccccccccc}
    \toprule
    Metric & Method & Cora & YelpChi & Reddit & T-Finance & Tolokers & Disney & Questions & Avg \\
    \midrule
    \multirow{3}{*}{AUC}
      & Full               & 0.6166 & 0.6152 & 0.5122 & 0.9114 & 0.5624 & 0.7246 & 0.5832 & \textbf{0.6465} \\
      & Sampling ($k=100$) & 0.6166$\pm$0.0000 & 0.6152$\pm$0.0000 & 0.5116 $\pm$ 0.0002 & 0.9019 $\pm$ 0.0013 & 0.5587 $\pm$ 0.0011 & 0.7246 $\pm$ 0.0000 & 0.5831$\pm$0.0000 & 0.6445 \\
      & Sampling ($k=50$)  & 0.6167$\pm$0.0001 & 0.6152$\pm$0.0000 & 0.5113 $\pm$ 0.0009 & 0.8927 $\pm$ 0.0016 & 0.5511 $\pm$ 0.0007 & 0.7246 $\pm$ 0.0000 & 0.5831$\pm$0.0000 & 0.6421 \\
    \midrule
    \multirow{3}{*}{AP}
      & Full               & 0.0684 & 0.0750 & 0.0356 & 0.6067 & 0.2459 & 0.1617 & 0.0400 & \textbf{0.1762} \\
      & Sampling ($k=100$) & 0.0684$\pm$0.0000 & 0.0750$\pm$0.0000 & 0.0356$\pm$0.0000 & 0.5888 $\pm$ 0.0031 & 0.2445 $\pm$ 0.0004 & 0.1617 $\pm$ 0.0000 & 0.0400$\pm$0.0000 & 0.1734 \\
      & Sampling ($k=50$)  & 0.0684$\pm$0.0000 & 0.0750$\pm$0.0000 & 0.0356 $\pm$ 0.0001 & 0.5619 $\pm$ 0.0017 & 0.2421 $\pm$ 0.0006 & 0.1617 $\pm$ 0.0000 & 0.0400$\pm$0.0000 & 0.1692 \\
    \bottomrule
    \end{tabular}
    }
\end{table*}

\begin{table*}[t]
    \centering
    \caption{End-to-end runtime comparison (in seconds) between exhaustive computation and the proposed sampling-based approximation.}
    \label{tab:efficiency_sampling}
    \resizebox{\textwidth}{!}{
    \begin{tabular}{c|ccccccc}
    \toprule
    Method & Cora & YelpChi & Reddit & T-Finance & Tolokers & Disney & Questions \\
    \midrule
    Full               & 3.23 & 2.32 & 2.57 & 2321.74 & 4.53 & 0.02 & 17.32 \\
    Sampling ($k=100$) & 3.29 & 2.32  & 1.54 & 36.61   & 2.12 & 0.02 & 9.17 \\
    Sampling ($k=50$)  & 2.68 & 2.29  & 1.46 & 32.58   & 1.99 & 0.02 & 8.39 \\
    \bottomrule
    \end{tabular}
    }
\end{table*}

\subsection{Effect of SVD Projection Dimension}
\label{app:svd-rank}

The proposed method projects raw node features into an $r$-dimensional space via 
truncated SVD before computing neighbor diversity. We fix $r{=}8$ throughout the 
main experiments and justify this choice by sweeping 
$r \in \{8, 16, 32, 64\}$ on test datasets whose raw feature dimension is large 
enough to support the full range: Cora ($d{=}1433$), YelpChi ($d{=}32$), Reddit 
($d{=}64$), and Questions ($d{=}301$). T-Finance ($d{=}10$), Tolokers ($d{=}10$), 
and Disney ($d{=}28$) are excluded because their raw dimensions preclude the upper 
end of the sweep.

Table~\ref{tab:svd-rank} reports the results. Three observations emerge. First, 
$r{=}8$ achieves the best average AUC (0.5818) and AP (0.0548) across the four 
datasets, confirming it as a robust cross-domain default. Second, the optimal $r$ 
varies per dataset: YelpChi favors $r{=}32$, reflecting that its raw dimension 
($d{=}32$) is already compact and benefits from retaining nearly all directional 
information, while Cora degrades sharply as $r$ grows (0.6166$\to$0.2437 AUC), 
because its high-dimensional raw features ($d{=}1433$) contain many low-energy 
noisy directions that dilute cosine similarity once included. Third, no single $r$ 
dominates on every dataset, the optimal projection rank depends on the intrinsic 
signal rank of each graph's attribute space, which is not available a priori in 
the zero-shot setting. We therefore fix $r{=}8$ as the default, which trades 
marginal per-dataset gains for cross-domain stability. The Average (Avg) column at $r{=}64$ 
is marked "---" because YelpChi's $d{=}32$ precludes $r{=}64$; averaging over a 
different set of datasets from the other rows would not be directly comparable.
\begin{table}[h]
\centering
\caption{Effect of SVD projection dimension $r$. Avg is computed across the four 
datasets shown. Bold indicates the best per column.}
\label{tab:svd-rank}
    \small
    \begin{tabular}{c|cccccc}
    \toprule
    Metric & $r$ & Cora & YelpChi & Reddit & Questions & Avg \\
    \midrule
    \multirow{4}{*}{AUC}
     & 8  & 0.6166 & 0.6152          & 0.5122 & 0.5832          & \textbf{0.5818} \\
     & 16 & 0.5343          & 0.6761          & 0.4684          & 0.5857          & 0.5661 \\
     & 32 & 0.3065          & 0.6900 & 0.4782          & 0.5854          & 0.5150 \\
     & 64 & 0.2437          & —               & 0.4738          & 0.5908 & —      \\
    \midrule
    \multirow{4}{*}{AP}
     & 8  & 0.0684 & 0.0750          & 0.0356 & 0.0400          & \textbf{0.0548} \\
     & 16 & 0.0559          & 0.0879          & 0.0311          & 0.0397          & 0.0537 \\
     & 32 & 0.0403          & 0.0913 & 0.0327          & 0.0398          & 0.0510 \\
     & 64 & 0.0351          & —               & 0.0318          & 0.0405 & —      \\
    \bottomrule
    \end{tabular}
\end{table}

\subsection{Low-degree Node Handling}
\label{app:low-degree-handling}
NeighborDiv is a second-order neighborhood statistic and therefore requires at least two neighbors
to form a valid neighbor pair. For nodes with \(d_i<2\), the diversity statistic \(D_i\) is undefined.
Our main implementation assigns such nodes a neutral standardized score \(s_i=0\). This choice
does not indicate that low-degree nodes are necessarily normal; rather, it reflects that their
second-order neighbor-to-neighbor organization cannot be measured by the proposed statistic.

To quantify the scope of this issue, Table~\ref{tab:low_degree_stats} reports the number of valid
nodes with \(d_i\ge2\), the valid-node ratio, and degree statistics for all test datasets. The prevalence
of low-degree nodes is dataset-dependent. It is negligible on dense graphs such as T-Finance
(99.8\% valid nodes), but substantial on sparse graphs such as Questions, where only 27.7\% of
nodes have at least two neighbors. This motivates an explicit sensitivity analysis of the sparse-node
fallback assignment.

We compare three evaluation settings in Table~\ref{tab:low_degree_ablation}:
\begin{itemize}
    \item \textbf{Zero fallback}: the default setting used in the main experiments, where nodes with
    \(d_i<2\) receive \(s_i=0\).
    \item \textbf{Median fallback}: nodes with \(d_i<2\) receive the median standardized score of
    valid nodes. This is another neutral assignment that is less tied to the mean of the standardized
    score distribution.
    \item \textbf{Valid-only}: evaluation is restricted to nodes with \(d_i\ge2\), i.e., the subset on
    which NeighborDiv is directly defined.
\end{itemize}

The default zero fallback achieves the best average AUC (0.6465) and AP (0.1762). Replacing it
with the median fallback yields comparable average AP (0.1740) and a moderate decrease in average
AUC (0.6343). The valid-only evaluation obtains a similar average AUC (0.6425) and AP (0.1756)
to the full-node default evaluation. These results indicate that the main conclusions are not driven
by the arbitrary treatment of low-degree nodes. Instead, the NeighborDiv signal remains effective on
the valid-node subset, and using a neutral fallback mainly serves to produce a complete ranking over
all nodes.

We emphasize that this analysis does not claim that low-degree anomalies are fully solved by the
current method. Nodes with \(d_i<2\) are outside the direct scope of a second-order neighbor-pair
statistic. Designing a principled degree-aware fallback for isolated or degree-one nodes is a natural
future direction, for example by combining NeighborDiv with first-order node-to-neighbor residual
scores or global feature outlier scores.
\begin{table}[t]
    \centering
    \caption{Low-degree node statistics on the seven test datasets. 
    \(V_{\mathrm{valid}}\) denotes nodes with \(d_i\ge2\), for which NeighborDiv is directly defined. 
    Invalid nodes are those with \(d_i<2\), which receive a neutral fallback score.}
    \label{tab:low_degree_stats}
    %\resizebox{\linewidth}{!}{
    \begin{tabular}{l|cccccc}
    \toprule
    Dataset & \(N\) & \(|V_{\mathrm{valid}}|\) & Valid ratio & Isol. & \(d=1\) & Invalid \((d<2)\) \\
    \midrule
    Cora      & 2,708  & 2,233  & 82.5\% & 0      & 475    & 475 \\
    YelpChi   & 23,831 & 15,679 & 65.8\% & 0      & 8,152  & 8,152 \\
    Reddit    & 10,984 & 8,332  & 75.9\% & 0      & 2,652  & 2,652 \\
    T-Finance & 39,357 & 39,275 & 99.8\% & 0      & 82     & 82 \\
    Tolokers  & 11,758 & 9,853  & 83.8\% & 930    & 975    & 1,905 \\
    Disney    & 124    & 113    & 91.1\% & 0      & 11     & 11 \\
    Questions & 48,921 & 13,567 & 27.7\% & 17,603 & 17,751 & 35,354 \\
    \midrule
    % Total / Avg & 137,683 & 89,052 & 64.7\% & 18,533 & 30,098 & 48,631 \\
    % \bottomrule
    \end{tabular}
    %}
\end{table}

\begin{table}[t]
\centering
\caption{Sensitivity analysis of low-degree node handling. Zero fallback is the default setting
used in the main experiments. Median fallback assigns nodes with \(d_i<2\) the median standardized
score of valid nodes. Valid-only evaluates only nodes with \(d_i\ge2\).}
\label{tab:low_degree_ablation}
\resizebox{\linewidth}{!}{
\begin{tabular}{c|ccccccccc}
\toprule
Metric & Setting & Cora & YelpChi & Reddit & T-Finance & Tolokers & Disney & Questions & Avg \\
\midrule
\multirow{3}{*}{AUC}
& Median fallback & 0.6315 & 0.5467 & 0.5122 & 0.9109 & 0.5656 & 0.6667 & 0.6067 & 0.6343 \\
& Zero fallback   & 0.6166 & 0.6152 & 0.5122 & 0.9114 & 0.5624 & 0.7246 & 0.5832 & 0.6465 \\
& Valid-only      & 0.6142 & 0.5164 & 0.5122 & 0.9122 & 0.5676 & 0.7477 & 0.6273 & 0.6425 \\
\midrule
\multirow{3}{*}{AP}
& Median fallback & 0.0704 & 0.0648 & 0.0356 & 0.6066 & 0.2477 & 0.1504 & 0.0427 & 0.1740 \\
& Zero fallback   & 0.0684 & 0.0750 & 0.0356 & 0.6067 & 0.2459 & 0.1617 & 0.0400 & 0.1762 \\
& Valid-only      & 0.0737 & 0.0270 & 0.0356 & 0.6084 & 0.2505 & 0.1851 & 0.0490 & 0.1756 \\
\bottomrule
\end{tabular}
}
\end{table}

\section{Controlled Homophily Sweep}
\label{app:homophily_sweep}

To systematically evaluate the sensitivity of our method to varying levels of graph homophily 
and to provide controlled evidence for the two-sided detection claim, we construct synthetic 
attributed graphs using the stochastic block model (SBM)~\cite{holland1983stochastic}.

\paragraph{Graph generation.} Each graph contains $n{=}2000$ nodes partitioned into $k{=}5$ 
equal-sized communities. The intra- and inter-community edge probabilities are calibrated to 
achieve a target edge homophily ratio $h \in \{0.1, 0.3, 0.5, 0.7, 0.9\}$ with an expected 
average degree of 15. Node features are generated as $x_i = \mu_{c_i} + \epsilon_i$, where 
$\mu_{c_i} \in \mathbb{R}^{50}$ is the community center sampled from $\mathcal{N}(0, 9I)$ 
and $\epsilon_i \sim \mathcal{N}(0, I)$. Features are preprocessed identically to the main 
experiments (truncated SVD to $r{=}8$ dimensions followed by $\ell_1$ row normalization).

\paragraph{Anomaly injection.} We inject 50 anomalous nodes per type by rewiring their edges:
\begin{itemize}
    \item \textbf{Type-H} (over-homogeneous): all neighbors are replaced with nodes from the \emph{same} community as the target node, creating an artificially concentrated neighborhood that deviates below the graph's typical diversity level.
    \item \textbf{Type-D} (over-diverse): neighbors are replaced with nodes sampled 
    \emph{uniformly across all} $k$ communities, creating a maximally dispersed neighborhood 
    that deviates above the typical diversity level.
    \item \textbf{Mixed}: both Type-H and Type-D anomalies are injected simultaneously 
    (50 each, 100 total, 5\% anomaly rate).
\end{itemize}

The edge homophily ratio $h$ controls the \emph{background} structure of normal nodes: at 
$h{=}0.9$, normal neighborhoods are already highly concentrated within their community, 
while at $h{=}0.1$, normal neighborhoods span multiple communities. The detectability of each anomaly type depends on its contrast against the background 
neighborhood organization. Type-H anomalies are most salient when normal neighborhoods 
are not already highly homogeneous. Type-D anomalies are expected to be salient when 
normal neighborhoods have a coherent community structure, but this contrast can break down 
in the extreme homophily regime, where the variance induced by mixed-community neighbors 
may overlap with the normal diversity range.

\paragraph{Results.}
Table~\ref{tab:homophily_sweep} reports AUC averaged over 3 random seeds. Three key observations emerge:
(1)~Our method achieves strong and robust detection of Type-H anomalies across all homophily levels (AUC $\ge 0.80$), substantially outperforming both baselines, which remain near random (AMEN-Ego: 0.52--0.56; NRS: 0.54--0.75).
(2) For Type-D anomalies, our method leads for $h \leq 0.7$ (AUC $\geq 0.78$).
At the extreme setting $h=0.9$, Type-D detection degrades for our method.
In this regime, normal neighborhoods already consist almost entirely of same-community nodes, 
so their pairwise similarities concentrate near a high value with small variance.
A Type-D anomaly whose neighbors span multiple communities induces a bimodal similarity distribution, 
but the induced variance can overlap with the normal diversity range, reducing the separation in 
the calibrated deviation $|D_i-D_{\mathrm{med}}|$.
First-order residual scores such as NRS are less affected in this regime because they aggregate 
neighbor features before comparison with the ego, making them complementary to the second-order 
neighbor diversity signal.
(3)~In the Mixed setting, our method maintains strong performance for $h \le 0.7$ (AUC $\ge 0.86$), as Type-H and Type-D signals jointly contribute to the overall anomaly ranking.
Overall, the results confirm that our two-sided scoring mechanism is effective across a wide homophily spectrum, covering the range of most real-world attributed graphs.

\begin{table}[h]
\centering
\caption{AUC under controlled homophily sweep on synthetic SBM graphs ($n{=}2000$, $k{=}5$). 
Type-H: over-homogeneous anomalies; Type-D: over-diverse anomalies; Mixed: both types combined.
Results averaged over 3 random seeds. Bold indicates best per setting.}
\label{tab:homophily_sweep}
\small
\begin{tabular}{llccccc}
\toprule
Anomaly Type & Method & $h{=}0.1$ & $h{=}0.3$ & $h{=}0.5$ & $h{=}0.7$ & $h{=}0.9$ \\
\midrule
\multirow{3}{*}{Type-H}
 & \textbf{NeighborDiv (ours)}     & \textbf{0.998} & \textbf{0.995} & \textbf{0.978} & \textbf{0.933} & \textbf{0.799} \\
 & NRS      & 0.752 & 0.700 & 0.677 & 0.634 & 0.537 \\
 & AMEN-Ego & 0.523 & 0.531 & 0.548 & 0.558 & 0.528 \\
\midrule
\multirow{3}{*}{Type-D}
 & \textbf{NeighborDiv (ours)}     & \textbf{0.785} & \textbf{0.778} & \textbf{0.880} & \textbf{0.847} & 0.362 \\
 & NRS      & 0.541 & 0.481 & 0.504 & 0.600 & \textbf{0.740} \\
 & AMEN-Ego & 0.594 & 0.626 & 0.580 & 0.530 & 0.645 \\
\midrule
\multirow{3}{*}{Mixed}
 & \textbf{NeighborDiv (ours)}     & \textbf{0.859} & \textbf{0.864} & \textbf{0.915} & \textbf{0.876} & 0.522 \\
 & NRS      & 0.619 & 0.580 & 0.588 & 0.620 & \textbf{0.654} \\
 & AMEN-Ego & 0.534 & 0.549 & 0.543 & 0.515 & 0.541 \\
\bottomrule
\end{tabular}
\end{table}

\section{Diagnostic Analysis on T-Finance}
\label{app:diagnostic}

Our method achieves the largest margin over baselines on T-Finance under the UMDT protocol (AUC $0.9114$ vs.\ the next-best $0.7828$ of GADAM, Table~\ref{tab:umdt_average}; AP $0.6067$ vs.\ $0.4842$, Table~\ref{tab:umdt_ap_appendix}). To better understand this unusually large margin and verify that NeighborDiv produces a meaningful
ranking on T-Finance, we provide four complementary post-hoc diagnostics: (i) the precision--recall curve under three sampling configurations, (ii) Precision@$K$ at multiple operating points, (iii) the score distribution separability between normal and anomalous nodes, and (iv) a comparison of the observed gap against the existing baseline variance reported in Table~\ref{tab:source_dependency}.

\begin{figure}[h]
\centering
\includegraphics[width=0.55\linewidth]{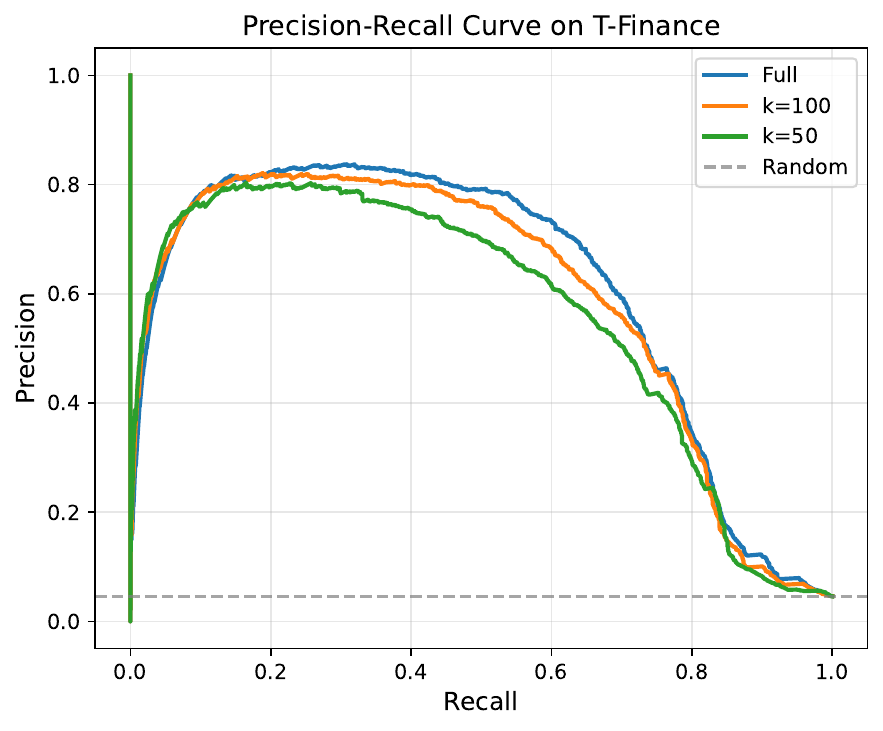}
\caption{Precision--recall curve of our method on T-Finance under three sampling configurations. The dashed gray line indicates the random-classifier baseline. NeighborDiv maintains high precision across a wide recall range, and sampled variants closely track exhaustive computation.}
\label{fig:tfin_pr}
\end{figure}

\paragraph{Precision--recall curve and top-$K$ precision.}
Figure~\ref{fig:tfin_pr} plots the precision--recall curve of our method on T-Finance under three sampling configurations (Full, $k{=}100$, $k{=}50$). The curve is smooth and remains substantially above the random-classifier baseline 
over a wide recall range. The three sampling configurations produce closely aligned 
curves (AP 0.6067, 0.5888, 0.5619), with Full dominating as expected. This indicates that the ranking quality is not sensitive to the sampling realization and that the reported AP is driven by a uniformly ordered score rather than a few extreme values. Table~\ref{tab:tfin_precision_k} further reports Precision@$K$ for $K \in \{100, 500, 1000, 5000\}$, where $K$ spans roughly $5.5\%$ to $277\%$ of the total anomaly count (1{,}803). Under the Full configuration, Precision@$500 = 0.830$ (i.e., $415$ of the top $500$ candidates are true anomalies), and Precision@$1000 = 0.804$, confirming that the top of the ranking is densely populated with genuine anomalies across multiple operating points rather than in only a narrow region.

\begin{table}[h]
\centering
\caption{Precision@$K$ on T-Finance under three sampling configurations. $K$ denotes the top-$K$ nodes ranked by anomaly score.}
\label{tab:tfin_precision_k}
\small
\begin{tabular}{lcccc}
\toprule
Configuration & $K{=}100$ & $K{=}500$ & $K{=}1000$ & $K{=}5000$ \\
\midrule
Full        & 0.580 & 0.830 & 0.804 & 0.295 \\
$k{=}100$   & 0.620 & 0.818 & 0.788 & 0.294 \\
$k{=}50$    & 0.620 & 0.800 & 0.747 & 0.289 \\
\bottomrule
\end{tabular}
\end{table}

\begin{figure}[h]
\centering
\includegraphics[width=0.55\linewidth]{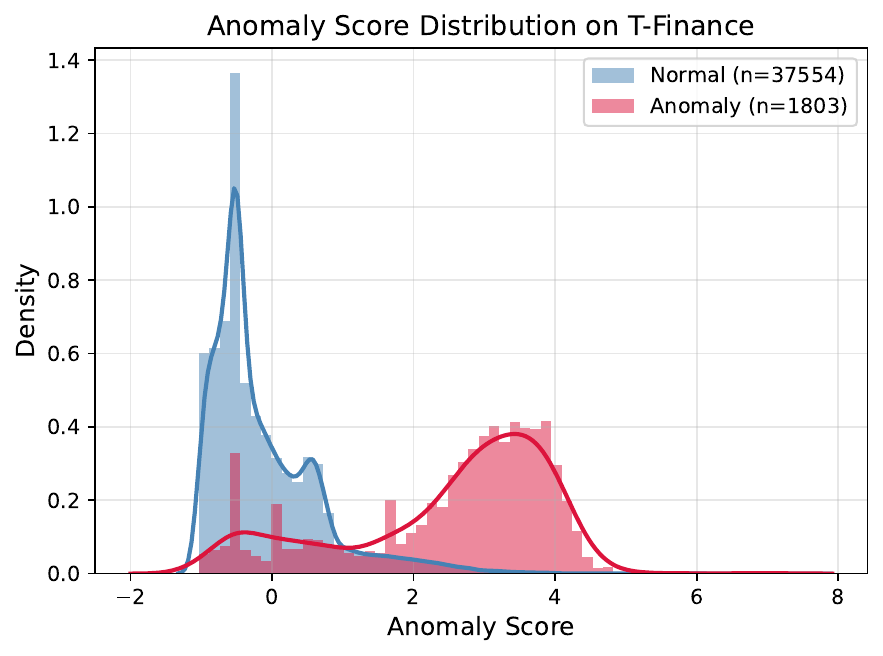}
\caption{Distribution of anomaly scores produced by our method on T-Finance, split by ground-truth label. Normal nodes (blue) concentrate sharply near zero with a narrow spread; anomalous nodes (red) form a broader distribution shifted to the positive side, with only marginal overlap around the normal mode. Normal and anomalous nodes form clearly separated score distributions, supporting neighbor diversity as an effective fraud-detection signal.}
\label{fig:tfin_score_dist}
\end{figure}

\paragraph{Score distribution separability.}
Figure~\ref{fig:tfin_score_dist} shows the distribution of our method's anomaly scores on T-Finance, split by ground-truth label. The two distributions are cleanly separated: normal nodes concentrate around a sharp mode near score $-0.5$ with mean $-0.121$ and standard deviation $0.794$, while anomalous nodes spread around a broad mode near score $3.5$ with mean $2.493$ and standard deviation $1.459$. The two means are separated by more than $3$ standard deviations of the normal group, and the Kolmogorov--Smirnov statistic between the two empirical distributions is $0.731$ with $p < 10^{-300}$ (below machine precision), indicating an extremely significant separation that cannot be explained by metric computation alone. This bimodal pattern is consistent with the mechanism argued in Section~\ref{subsec:neighbor_diversity}: fraudulent accounts in T-Finance tend to connect to an unusually heterogeneous set of users, producing neighbor diversity values far from the graph-level median and thus large calibrated deviations.

\paragraph{Baseline variance from existing source-domain analysis.}
While we do not re-run the learning-based baselines with additional random seeds (as several baselines involve non-trivial training pipelines that would require code modification), Table~\ref{tab:source_dependency} already provides a direct quantification of baseline fluctuation on T-Finance: the standard deviation of each generalist baseline's AUC across the four different auxiliary training sources under the SDIT protocol. On T-Finance, this source-level standard deviation is $0.0519$ for UNPrompt, $0.0959$ for AnomalyGFM, and $0.0448$ for IA-GGAD. The AUC gap of $0.1286$ between our method ($0.9114$) and the best baseline ($0.7828$) is therefore more than \(1.34\times\) times larger than the largest of these standard deviations. Since source-domain variation is typically a larger source of fluctuation than random-seed variation in training-based GGAD methods~\cite{niu2024zero,qiao2025anomalygfm}, the observed gap is clearly outside any plausible range of baseline variability. Our method itself is training-free and deterministic (std exactly zero, Table~\ref{tab:source_dependency}), so its own variance does not contribute to the gap.

\paragraph{Summary.}
The four diagnostics jointly support the reliability of the T-Finance result. \textbf{(i)} The precision--recall curve is smooth and consistent across sampling configurations, with Precision@$500 = 0.830$; \textbf{(ii)} Precision@$K$ remains high across multiple operating points, ruling out the possibility that the AP gain is concentrated in a single narrow region of the ranking; \textbf{(iii)} the score distribution exhibits a statistically overwhelming separation between normal and anomalous nodes ($\mathrm{KS}=0.731$, $p<10^{-300}$); and \textbf{(iv)} the observed AUC gap is larger than the measured source-domain standard deviations of the
generalist baselines under SDIT, although this comparison is only a proxy for baseline variability. The strong performance on T-Finance is therefore attributable to the suitability of neighbor diversity as an anomaly signal for this particular financial-fraud domain, rather than to a metric artifact or baseline randomness.

%%%%%%%%%%%%%%%%%%%%%%%%%%%%%%%%%%%%%%%%%%%%%%%%%%%%%%%%%%%%

%\newpage
%\input{checklist.tex}

\end{document}